%% file: acm_main.tex
\documentclass[sigconf]{acmart}

\usepackage{amsmath}
\usepackage{subcaption}
\usepackage[ruled, linesnumbered]{algorithm2e}

\AtBeginDocument{%
  }

\setcopyright{acmcopyright}
\copyrightyear{2024}
\acmYear{2024}
\acmDOI{XXXXXXX.XXXXXXX}

\acmConference[XXXX'25]{Make sure to enter the correct
  conference title from your rights confirmation email}{XXX - XXXX,
  2025}{XXX}

\begin{document}


\title{State Frequency Estimation for Anomaly Detection}



\author{Clinton Cao}
\orcid{1234-5678-9012}
\affiliation{%
  \institution{Delft University of Technology}
  \city{Delft}
  \country{The Netherlands}
}

\author{Agathe Blaise}
\orcid{1234-5678-9012}
\affiliation{%
  \institution{Thales SIX GTS France}
  \city{Gennevilliers}
  \country{France}
}

\author{Annibale Panichella}
\orcid{1234-5678-9012}
\affiliation{%
  \institution{Delft University of Technology}
  \city{Delft}
  \country{The Netherlands}
}

\author{Sicco Verwer}
\orcid{1234-5678-9012}
\affiliation{%
  \institution{Delft University of Technology}
  \city{Delft}
  \country{The Netherlands}
}


\renewcommand{\shortauthors}{Cao et al.}

\begin{abstract}
Many works have studied the efficacy of state machines for detecting anomalies within NetFlows. These works typically learn a model from unlabeled data and compute anomaly scores for arbitrary traces based on their likelihood of occurrence or how well they fit within the model. However, these methods do not dynamically adapt their scores based on the traces seen at test time. This becomes a problem when an adversary produces seemingly common traces in their attack, causing the model to miss the detection by assigning low anomaly scores. We propose SEQUENT, a new unsupervised approach that uses the state visit frequency of a state machine to adapt its scoring dynamically for anomaly detection. SEQUENT subsequently uses the scores to generate root causes for anomalies. These allow the grouping of alarms and simplify the analysis of anomalies. We evaluate SEQUENT's effectiveness in detecting network anomalies on three publicly available NetFlow datasets and compare its performance against various existing unsupervised anomaly detection methods. Our evaluation shows promising results for using the state visit frequency of a state machine to detect network anomalies.
\end{abstract}





\maketitle

\input{sec_introduction}
\input{sec_preliminaries}
\input{sec_threat_model}
\input{sec_related_work}
\input{sec_methodology}
\input{sec_setup}

\input{sec_results}
\input{sec_conclusion_future}

\section*{Acknowledgments}
This work is funded under the Assurance and certification in secure
Multi-party Open Software and Services (AssureMOSS) Project,
(https://assuremoss.eu/en/), with the support of the European Commission and H2020 Program, under Grant Agreement No. 952647.
\bibliographystyle{ACM-Reference-Format}
\bibliography{references}

\end{document}

%% file: sec_introduction.tex
\section{Introduction}\label{sect-introduction}
NetFlow encapsulates summarized network statistics of a connection between two arbitrary hosts and can be used for traffic analysis. It is a popular alternative to storing individual packets collected from a network due to its benefits: easier to collect, simpler to process, and less intrusive to the privacy of the network users as there is no possibility to inspect every packet. Despite these benefits, detecting network anomalies (e.g., network attacks) within NetFlows (or simply flows) remains a challenge, as essential information might be lost within the summarized statistics. Various works have studied the use of supervised and unsupervised machine learning (ML) models to detect network anomalies within NetFlow data~\cite{Nguyen2019_gee,Zoppi2021_unsupervised,Venturi_2021_Drelab,Fosic_2023_Anomaly,Waseem_2024_Real, Zoppi2021_meta_learning,Campazas_2023_Malicious}. 

While there has been success in applying ML models to detect anomalies in NetFlow data, a challenge faced by many security analysts is the interpretability of the predictions made by the ML model~\cite{nadeem_2023_sok}. Understanding these predictions enables analysts to determine whether the model has made the correct decision and to gain confidence in the ML model deployed in the ADS.

To address this challenge, several studies have explored the use of state machines to detect network anomalies within NetFlow data, as these models are considered inherently interpretable~\cite{Pellegrino2017_learning, cao_learning_state_machines22, Grov19_towards}. Typically, these studies train the model using traces generated from benign flows and compute an anomaly score based on the likelihood of those traces according to the model.

Although a standard approach in ML, we argue that this has two limitations, making it not well-suited for detecting anomalies we see in practice. Firstly, many benign flows occur with a low probability, resulting in many false alarms. Secondly, malicious flows tend to resemble common flows, giving them large probabilities and thus making it hard to raise alarms. To address this, we introduce SEQUENT (short for \textbf{S}tat\textbf{e} Fre\textbf{qu}ency \textbf{E}stimation for A\textbf{n}omaly De\textbf{t}ection) to overcome these limitations by using a key feature of the state machine model: its behavioral structure. Instead of computing likelihoods, we compute visit frequencies of different states within the model. Each state essentially represents a distinct type of network traffic behavior (a cluster or latent value). When we observe behavior that occurs much more frequently than expected, an alarm is raised. Intuitively, benign flows with low probability do not trigger alarms if they occur at the expected frequency. Moreover, malicious flows with high probability do raise alarms when they occur more frequently than expected. Note that this also detects infrequent malicious flows not observed at training since their expected count is assumed to be very small. The drawback is that normal flows may raise alarms when they occur more frequently. While this can happen, our experiments demonstrate that these flows have a smaller impact compared to the malicious flows.  

Similar to previous studies that utilized state machines, SEQUENT learns a state machine from benign flows to model sequential network behavior. The process begins by discretizing the flow features into a symbolic representation. To achieve this, we developed a discretization method~\cite{encode_repo} inspired by the GloVe algorithm~\cite{Pennington_2014_Glove}. Each selected feature is discretized independently and each flow is mapped to a symbol, created by concatenating the discretized feature values of the corresponding flow. 

SEQUENT then runs each trace observed at test time through the learned model and computes the set of states visited by the trace. It uses the computed sets to update the visit frequency of each state in the model. SEQUENT compares the visit frequencies to the expected frequencies observed during training and raises an alarm whenever a behavior occurs more frequently than expected. 
 
Another well-known challenge faced by many security analysts in their daily routine is the large volume of alerts raised by the anomaly detection system (ADS)~\cite{nadeem2022_alert,lanvin2023_towards,bushra2022_99false}. The sheer volume of alerts makes it difficult for analysts to prioritize which anomalies to inspect first, turning the inspection of alerts into a labor-intensive task. 
SEQUENT addresses this issue by computing root causes (state identifiers) for each detected anomaly, which allows grouping and ranking of anomalies (e.g., using the frequency of a root cause). Additionally, SEQUENT establishes links between the root causes and their corresponding flows. The aim is to provide analysts with concrete instances of anomalous behavior, enabling a quick and in-depth analysis of anomalies. 

We empirically demonstrate SEQUENT's effectiveness in the task of detecting various types of network anomalies using three publicly available NetFlow datasets~\cite{assuremoss_dataset, Garcia2014_an_empirical, Macia-Fernandez2018_ugr}. Our experiment results demonstrate that SEQUENT is highly effective in the detection of network anomalies, outperforming existing state-of-the-art unsupervised methods. 
The main contributions of this paper are summarized as:

\begin{itemize}
    \item We present SEQUENT, a novel approach that employs state visit frequencies of a state machine for anomaly detection.
    \item SEQUENT offers the capability to group and rank anomalies, empowering analysts to filter and prioritize which anomalies they should investigate.
    \item We compare SEQUENT against existing state-of-the-art unsupervised methods on three different datasets, demonstrating its effectiveness and generalizability.  
\end{itemize}


%% file: sec_preliminaries.tex
\section{Preliminaries}\label{sec:preliminaries}

\subsection{State Machine}
A state machine, formally known as a finite state machine or finite state automaton, is a mathematical model used for modeling the sequential behavior of a system. It is formally defined as a 5-tuple $(\Sigma, Q, q_0, \delta, F)$\cite{sipser2013_introduction}, where:

\begin{itemize}
    \item $\Sigma$ is a finite alphabet.
    \item $Q$ is a finite set of states.
    \item $q_0 \in Q$ is a unique start state.
    \item $\delta : Q \times \Sigma \to Q$ is the transition function that determines which transition should be taken based on the current state and the (new) symbol read from the input.
    \item $F \subseteq Q$ is the set of final states.
\end{itemize}

\begin{figure}
    \centering
    \includegraphics[width=\columnwidth]{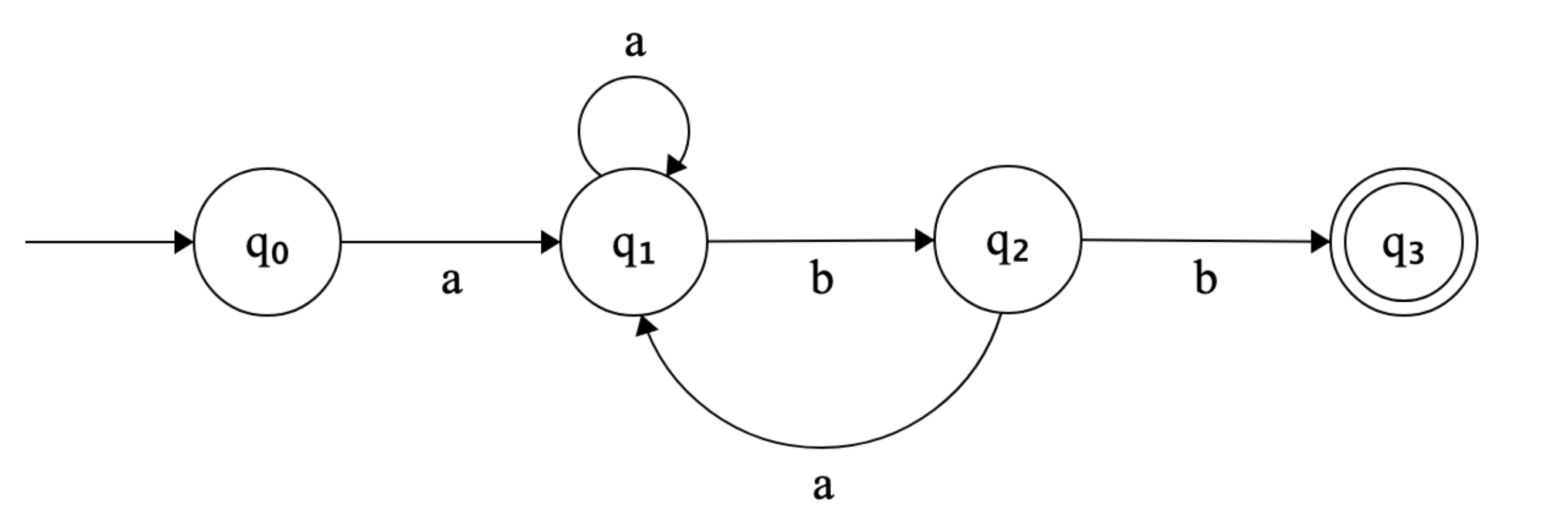}
    \caption{An illustrative example of a state machine. At the end of the input trace, $aabab$, we would end up in state $q_2$.}
    \label{fig:example_state_machine}
\end{figure}

Given an input trace $t$~(a sequence of events, where each event $e \in \Sigma$) and a system $S$, one can use the state machine to determine in which state $S$ would end up, thereby understanding the behavior the system would exhibit. In the context of this work, each event in a trace represents a single flow. An illustrative example of a state machine is shown in Figure~\ref{fig:example_state_machine}. The state machine in this example consists of only one final state, $q_3$.

\subsection{State Machines Considered in SEQUENT}
State machines come in two flavors, namely deterministic or non-deterministic state machines~\cite{sipser2013_introduction}. SEQUENT solely considers deterministic state machines as they are easier to interpret; simpler to predict the next event, and verifying the same sequence of events would execute the same sequential behavior is more straightforward~\cite{Lee_2021_determinism}. With a non-deterministic state machine, multiple transitions can be triggered simultaneously, even when no event has occurred. These non-deterministic transitions can inaccurately increase the visit frequencies of states within a state machine, causing states (or behaviors) to be falsely flagged as anomalous. Since SEQUENT aims to address the challenge of large volume alerts by providing root causes for anomalies, non-deterministic transitions can result in providing false root causes to analysts. 

Furthermore, the use of state machine is intended provide analysts with a way to inspect the sequential behaviors exhibited by an arbitrary system. However, when multiple transitions are triggered simultaneously within a non-deterministic state machine, it becomes difficult for an analyst to determine which path should be followed in the model given an arbitrary trace. For these reasons, SEQUENT focuses only on learning deterministic state machines.

\begin{figure}
    \centering
    \includegraphics[scale=0.65]{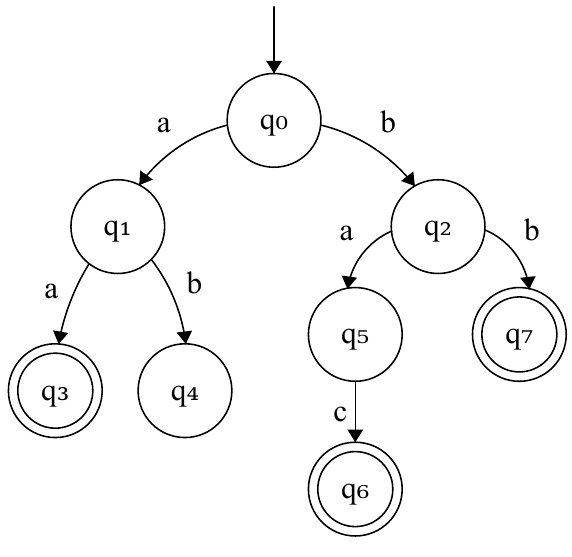}
    \caption{An example of a PTA.}
    \label{fig:example_pta}
\end{figure}

We can employ two approaches for learning a state machine from trace data: active and passive learning. We opt for the latter approach as it only requires us to collect traces from the target system $S$ instead of continuously interacting with $S$ to learn a state machine~\cite{Angluin_1987_learning}. State-merging heuristic algorithm is a highly effective method for learning state machines in the passive-learning approach~\cite{lang_1998_results}. It begins with a tree-like state machine, known as the Prefix Tree Acceptor (PTA) (see Figure~\ref{fig:example_pta}). It then iteratively compares pairs of states to identify states that can be merged. Merges are done based on the scores computed during the comparisons; for example, states are merged when their merge score has exceeded the minimum score threshold. The algorithm ends when it cannot find any new merges. The final output is an approximation of the smallest state machine consistent with the input data.

%% file: sec_threat_model.tex
\section{Assumptions of the Adversary}\label{sec:adversary_assumptions}
Our work is grounded in a set of assumptions we formulated concerning the objectives and capabilities of an adversary. We consider an adversary attacking an arbitrary target network $TN$, driven by various objectives such as infiltrating $TN$ to find machines containing sensitive data, disrupting normal operations by launching denial of service attacks against specific services, orchestrating a botnet, or infecting machines with malware. In our main experiments, the training and test data from the datasets simulate an adversary attacking $TN$ without taking our defense into account. In our robustness experiments, we assume a more adept adversary who is aware of an ADS within $TN$ but unaware of its specifics. Their goal is to remain undetected until they have accomplished their objectives. To accomplish this, the adversary can monitor $TN$ to collect normal network behaviors and then attempt to bypass the ADS by generating behaviors resembling those collected during normal operations within $TN$. The adversary can only listen to the data that was sent but cannot delete or alter any of the sent data.

%% file: sec_related_work.tex
\section{Related Works}\label{sec:related_works}
\textit{\textbf{Traditional Methods.}} Several works have studied the efficacy of unsupervised learning for network anomaly detection. The primary focus has been on utilizing traditional unsupervised ML algorithms (e.g., Isolation Forest). Zoppi et al. conducted multiple studies evaluating the effectiveness of traditional unsupervised methods for anomaly detection on NetFlow data~\cite{Zoppi2021_unsupervised,Zoppi2021_meta_learning}. Campazas-Vega et al. also studied the effectiveness of Isolation Forests in detecting anomalies within NetFlow data. In contrast to SEQUENT, these studies do not dynamically adapt the anomaly scores based on the events/traces seen at test time for anomaly detection.

\textit{\textbf{State Machines.}} Numerous studies have investigated the efficacy of state machines for anomaly detection. Pellegrino et al. introduced BASTA, a system that learns probabilistic timed state machines from NetFlows and uses them as behavioral fingerprints for machines infected with botnet malware~\cite{Pellegrino2017_learning}. Alerts are raised when traces match behaviors captured in the fingerprints. Unlike SEQUENT, BASTA uses malicious data for learning rather than solely relying on benign data. Additionally, BASTA learns a state machine for each malware. This approach requires analysts to maintain an updated list of malware samples to learn fingerprints. Any missing malware samples may result in the system failing to detect new or unknown malware. SEQUENT learns a single state machine that models the behavior exhibited in a network and is not specific to any particular malware or network anomalies.

Lin et al., Cao et al.,  and Matoušek et al. learned probabilistic state machines to detect anomalies in different systems\cite{cao_learning_state_machines22, Lin18_TABOR, Matoursek_21_automata}. These works raise alerts when a new trace deviates significantly from what was observed during training. The degree of deviation is quantified based on the likelihood of a trace. In contrast, SEQUENT quantifies the abnormality based on state visit frequency for a corresponding trace and compares this to what was observed during training. 

Grov et al. presented a different approach for learning state machines to detect anomalies in NetFlows~\cite{Grov19_towards}. The approach learns state machines using solely traces of network protocols extracted from benign data. The degree of abnormality is based on the normalized edit distance of an arbitrary trace under scrutiny; the number of changes needed such that each trace event can be replayed in the state machine. In contrast to SEQUENT, this work does not utilize the frequency of the states to adapt the scores dynamically at test time. Furthermore, it solely considers network protocols for the learning, whereas SEQUENT considers other features.

\textit{\textbf{Neural Networks.}} Deep Neural Networks (DNN) gained increasing popularity among researchers in recent years due to their ability to identify complex hidden patterns that traditional models may struggle to detect. Mirsky et al. introduced Kitsune, a state-of-the-art solution that utilizes an ensemble of Autoencoders (AE) to detect network anomalies in an unsupervised manner. It extracts various statistics from benign data and uses them for training. Kitsune then attempts to recreate instances it has seen during training by computing similar statistics for arbitrary new data. Alarms are raised for instances with large recreation errors\cite{Mirsky_2018_kitsune}. Neural networks are notoriously known to be difficult to interpret, as it is not always clear what happens between the input and output layers. Nguyen et al. sought to address this issue with their framework, GEE. It uses a Variational Autoencoder (VAE) to detect anomalies and generates explanations by providing SHAP-like plots of the gradients computed for NetFlow features\cite{Nguyen2019_gee}. However, it is not clear how a security analyst should interpret these gradients. In contrast, SEQUENT learns an inherently interpretable model and generates root causes, which can be used to facilitate the understanding of (malicious) sequential behaviors. Root causes are linked back to concrete instances producing the anomalous behaviors.

%% file: sec_methodology.tex
\section{SEQUENT}\label{sec:SEQUENT}

\subsection{Intuition Behind SEQUENT}\label{subsec:intuition_SEQUENT}
SEQUENT employs a novel approach for computing the anomaly score of an arbitrary trace under scrutiny. Unlike the measures of abnormality presented in prior works~\cite{Pellegrino2017_learning,Matoursek_21_automata, Lin18_TABOR, Grov19_towards, cao_learning_state_machines22}, SEQUENT leverages the frequencies of the state visits in a state machine to compute an adaptive anomaly score. The underlying principle of SEQUENT is that behaviors occurring significantly more often than those observed during training are indications of anomalies. Traces under scrutiny are replayed on a state machine to determine how well they fit within the learned model. Each trace visits a set of states, and the frequency of visits to each state is compared to the expected frequency observed during training. When the frequency of visits to a state is significantly higher than its expected frequency, it signals that unusual behavior is occurring in this state. The more significant the increase in visits for this state, the more likely it is to show anomalous behavior. SEQUENT employs this rationale to dynamically adapt its scores for arbitrary traces seen at test time. SEQUENT's intuition will also work for (anomalous) behaviors never seen before during training, as these will visit unusual paths during the replay.  

\subsection{SEQUENT's Anomaly Detection Pipeline}\label{subsec:SEQUENT_pipeline}
\begin{figure*}[t]
    \centering
    \includegraphics[scale=0.45]{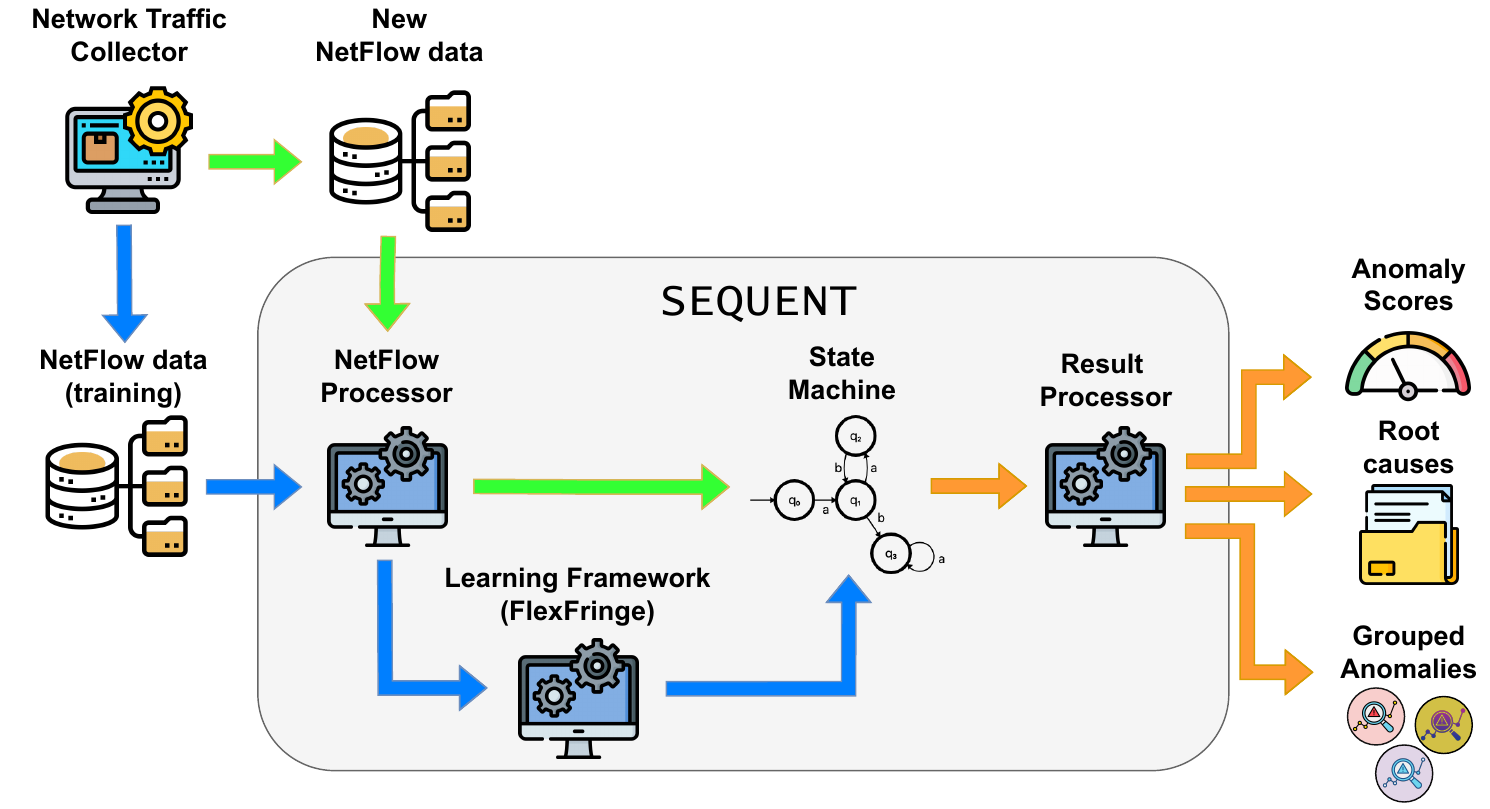}
    \caption{SEQUENT's anomaly detection pipeline.}
    \label{fig:SEQUENT_pipeline}
\end{figure*}

Figure~\ref{fig:SEQUENT_pipeline} depicts SEQUENT's anomaly detection pipeline. It provides a high-level overview illustrating the flow of learning of a state machine from training data (represented by the blue arrows), the flow for running predictions on new data (represented by green arrows), and the flow for generating the results (represented by the orange arrows). Before NetFlows are used for training or prediction, they are first transformed into traces by the NetFlow Processor component, as depicted in the figure. Each flow is described in the following paragraphs.

\textit{\textbf{Processing NetFlows.}}\label{subsec:process_netflows}
NetFlow features need to be discretized before learning a state machine, given that each flow contains both continuous and discrete feature data. Furthermore, each flow needs to be transformed into an event (discrete symbol) to generate traces (sequence of events). Although
the discretization process is not strictly necessary for the learning of a state machine, using using the raw NetFlow feature data (particularly continuous data) will lead to the combinatorial explosion of states. Discretization enables us to reduce the number of possible combinations in the feature data, thereby also reducing the number of states used in the state machine. 
To discretize NetFlow features, we develop an encoding algorithm inspired by the GloVe algorithm~\cite{Pennington_2014_Glove}. Our encoding algorithm computes co-occurrences between values for a given input flow feature $f$ and stores them in a matrix. Each row denotes a unique value $v$ of $f$ and how often it occurs with another value $v'$. Feature values are discretized by first clustering the rows using KMeans and then using cluster assignments as their discretization label. We make our encoding algorithm available on Github~\cite{encode_repo}. 

To transform a flow into an event symbol, we concatenate the discretized feature values of the selected features with an underscore symbol to form a single event symbol. For example, if three features are used to represent an event, we get the following event symbol: $DFV1\_DFV2\_DFV3$, where $DFV$ is the abbreviation for discretized feature value. In the case of one feature (e.g., bytes), we directly use its discretized values. To convert events into traces, we apply a sliding window of length $N$ to generate traces. These are then subsequently used to learn a state machine. Traces that do not have a length of $N$ are excluded from learning as they do not provide sufficient information.

\textit{\textbf{Learning of a State Machine.}}\label{subsec:learning_from_netflow}
SEQUENT utilizes FlexFringe~\cite{Verwer2017_Flexfringe} to learn a state machine from NetFlow traces. This framework offers, out of the box, various heuristics for learning a state machine. Additionally, it enables its users to learn other types of models such as Markov Chains. FlexFringe also automates the creation of traces from the CSV file provided as input data. The user has to specify which features to use to create events, the length of the sliding window, and how the data should be sorted (e.g., by connections) to create the traces. 

Since Flexfringe's trace creation feature of FlexFrige matches the trace creation method mentioned in the previous paragraph, we leverage this feature of FlexFringe to generate traces from the provided input NetFlows. We opt for FlexFringe in the design of SEQUENT based on these aforementioned features.

Once the selected features of the training data are discretized, we forward the CSV containing discretized feature values of the training to FlexFringe to create traces and learn a state machine. We specify to FlexFringe to use the discretized features values for the creation of the event symbols.

When learning is completed, FlexFringe outputs a state machine along with various statistics, such as how often each state of the state machine was visited during training. We store the model and the statistics for the computation of the rolling anomaly score for traces observed at test time. 
 
\textit{\textbf{Computing Rolling Anomaly Scores.}}\label{subsec:rolling_score}
At test time, the new NetFlow data is processed in the similar manner as with the training data and is forwarded to FlexFringe. FlexFringe simulates each trace generated from the test data and computes the sequence of states visited by each trace. For each computed state sequence, we compute its rolling anomaly scores with the help of the statistics extracted during training of the model. Algorithm~\ref{algo:sequent_rolling_score} depicts the pseudocode for computing rolling anomaly scores for an arbitrary trace $t$. 

The first step in computing the rolling anomaly scores is to extract the state visit frequencies of each state and the sum of all visit frequencies (see lines 4 to 8). The sum of all visit frequencies (stored in ``$TotalVisits$" from line 3) is used for normalization, as each state may have been visited a different number of times during training, causing some states to be more frequent than others. Note that lines 2 to 8 do not have to be repeated for each arbitrary trace we observe at test time. However, we have included it in Algorithm~\ref{algo:sequent_rolling_score} for the sake of completeness. In our implementation of SEQUENT, we compute this once and store the result in memory. 

The next step in computing rolling anomaly scores is to simulate the (test) trace $t$ on the state machine and extract the path it traverses in the model (line 10). For each state $q'$ in the extracted state sequence (i.e., the path), we compute its anomaly score (lines 11 to 18). Note that since we are working with visit frequencies, large counts can have a considerable impact on the score. To address this, we use the logarithm of the frequencies to compute the score for $q'$ (as shown on lines 12 to 17). The addition of one within the logarithm is used to ensure that we do not encounter the undefined behavior when of calculating the logarithm of zero. The base of the anomaly score for $q'$ starts with its visit frequency observed so far at test time (line 12). If the $q'$ has not been observed yet, a value of zero will be returned. The observed frequency (at test time) is then compared to the corresponding frequency extracted from the training statistics (line 13) and the score is normalized (lines 14 and 15). Finally, the visit frequency of state $q'$ is updated (line 16), and the score is added to the results list (line 17). The anomaly score of trace $t$ is the sum of all individual scores in the result list. We keep the individual scores, as they are used to compute the root cause of an anomaly (explained in the next paragraph).    

\begin{algorithm}
\LinesNumbered
\caption{SEQUENT's Rolling Anomaly Score}\label{algo:sequent_rolling_score}
\KwIn{State Machine $S$, Trace $t$, Training Statistics $TS$}
\KwOut{Rolling anomaly scores of $t$}
$AnomalyScores \leftarrow []$\\
$StateFreq_{train} \leftarrow \emptyset$\\
$TotalVisits \leftarrow 0$\\ 

\For{ each state $q$ in $S$} {
 $Freq_q \leftarrow extractVisitFreq(TS, q)$\\
 $StateFreq_{train}[$q$] \leftarrow freq_{q}$\\
 $TotalVisit \leftarrow TotalVisit + Freq_q $\\
}

$StateFreq_{test} \leftarrow \emptyset$\\
$StateSequence \leftarrow computeStateSequence(S, t)$\\
\For{each state $q'$ in $StateSequence$} {
  $StateScore \leftarrow log(StateFreq_{test}[$q'$] + 1)$\\
  $StateScore \leftarrow StateScore - log(StateFreq_{train}[$q'$] + 1)$\\
  $StateScore \leftarrow StateScore + log(TotalVisit + 1)$\\
  $StateScore \leftarrow StateScore + log(|StateFreq_{test}| + 1)$\\
  $StateSeq_{test}[$q'$] \leftarrow StateSeq_{test}[$q'$] + 1$\\  
  $AnomalyScores.add(StateScore)$\\
}

\Return AnomalyScores
\end{algorithm}

\textit{\textbf{Computing Root Causes.}}
An advantage of using a state machine to model sequential behavior is the ability to follow and analyze the path that an arbitrary trace $t$ takes within the model. We leverage this capability to compute a root-cause symbol for $t$. We do so by finding the state with the largest rolling anomaly score visited by $t$ and using the state identifier as the root-cause symbol. It denotes the state that causes the largest anomalous activity in $t$. As an example, let us use the model shown in Figure~\ref{fig:example_state_machine} as the model that was learned from benign flows. Let $t{=}[a, a, b, a, b, a, a]$ be the trace under scrutiny, $seq{=}[1, 1, 2, 1, 2, 1, 1]$ be the sequence of states visited by $t$ and $a {=} [0.19, 0.25, 0.18, 0.30, 0.20, 0.35, 0.40]$ be the rolling anomaly scores computed for $seq$. State 1 has the highest anomaly score (0.40), making ``1" the root-cause symbol for $t$. 

\textit{\textbf{Grouping Anomalies.}}
A key benefit of SEQUENT is the ability to create groups of anomalies exhibiting similar network behavior. SEQUENT leverages the root-cause symbols to group anomalies. The rationale is that traces sharing the same root-cause symbol denote behaviors exhibiting a notable level of anomalous activity in the same state. As traces signify the paths taken within the state machine, and the anomaly score of a state increases based on its visit frequency, traces sharing the same root-cause symbol will frequently visit this particular state. These traces will partly consist of the same events.   

\textit{\textbf{Linking Root Cause Back to Data.}}
Another benefit of SEQUENT is its ability to link root causes to the corresponding flow, producing a notable level of anomalous activity. SEQUENT achieves this by utilizing the insights produced by FlexFringe after making predictions on a set of arbitrary traces. For each prediction of a trace $t$, FlexFringe reports details on which lines of the CSV file were used for the creation of $t$. Specifically, the start and end points are reported. SEQUENT utilizes this information to pinpoint the exact flow within the CSV file, producing a notable level of anomalous activity in $t$ and links this to the root-cause symbol. This enables us to collect concrete instances of NetFlows for anomalies that are grouped under the same root-cause symbol.

%% file: sec_setup.tex
\section{Experiment Setup}\label{sec:experiment_setup}

\subsection{Dataset Selection}
To showcase SEQUENT's effectiveness in identifying network anomalies in an unsupervised setting, we have selected three publicly accessible datasets. These datasets comprise malicious NetFlows originating from a wide range of attacks. This allows us to assess SEQUENT's ability to identify various types of attacks. Furthermore, the varied nature of these datasets enables us to evaluate SEQUENT's generalizability across different network architectures.

\textbf{\textit{AssureMOSS Dataset.}} The first dataset consists of NetFlows collected from a Kubernetes cluster\cite{assuremoss_dataset} and used in the study conducted by Cao et al.~\cite{cao_learning_state_machines22}. It contains benign flows generated by real-life users and malicious flows generated by three attack scenarios. The three scenarios are constructed using a specialized threat matrix of the MITRE ATT\&CK framework, designed explicitly for Kubernetes~\cite{cao_learning_state_machines22}. Each attack is fragmented into various steps that an adversary has to execute (e.g., initial access, execution, discovery, etc.). The three attack scenarios cover attacks such as (1) exploiting a vulnerability to discover secrets and using them to do privilege escalation, (2) executing a denial of service (DoS) attack, and (3) exploiting a vulnerability that allows the deployment of malware. Cao et al. created the dataset to investigate the effectiveness of state machines for the aforementioned network attacks. For this reason, we selected this dataset as one of the baselines for our work.

\textbf{\textit{CTU-13 Dataset.}} The second dataset is a well-known NetFlow dataset consisting of a large collection of realistic botnet network traffic mixed with normal and background traffic. Specifically, there are thirteen captures of various botnet samples. Each comprises various actions and protocols exhibited by the deployed malware~\cite{Garcia2014_an_empirical}. The captures cover a wide range of attacks, e.g., distributed DoS, port scans, Click-Fraud, and sending spam. The benign and background flows were collected from normal machines deployed on a university network, and malicious flows were collected from virtual machines that were running the different botnets. Each virtual machine was bridged onto the university network to collect flows for the final dataset. This dataset has been used in multiple prior studies that investigated the effectiveness of state machines in detecting botnet traffic (e.g., Pellegrino et al.~\cite{Pellegrino2017_learning} and Grov et al.~\cite{Grov19_towards}) and serves as another suitable baseline for our work. 

\textbf{\textit{UGR-16 Dataset}}. The third and final dataset consists of 100 GB of NetFlows collected from a Spanish ISP. The benign flow data was collected from real-life clients of the ISP, and the malicious flow data was collected from five machines that were used to launch the attacks. The attacks were not launched against any clients but rather five machines that were set up as victim machines~\cite{Macia-Fernandez2018_ugr}. The dataset mostly covers various types of DoS and scanning attacks. Furthermore, it also covers attacks in which spam emails are sent and botnet traffic. The botnet traffic was collected from the CTU-13 dataset and was modified before they were inserted into this dataset. The modifications consist of matching the IP addresses and timestamps of the flows to the ones used in their setup. We selected this dataset as one of our baseline datasets because it contains a wide range of network behaviors produced by various users.


\subsection{Creating Train/Test Splits}\label{subsec:splitting_data}
We outline the procedure for creating the train/test split for each dataset. Before the split is created, NetFlows are first sorted by timestamp and then by connections. We believe connection data has less noise than data sorted by timestamps, and this would help with learning more accurate models. 

The AssureMOSS dataset was collected over a two-hour period, during which no attack scenarios were initiated in the first hour. In the study conducted by Cao et al., the first hour was used as the training data because it contains only benign flows, while the second hour was used as the test data. Because SEQUENT also learns a state machine from benign flows for anomaly detection, we adopt the same train/test split as used by Cao et al. This results in a training set containing 120,000 flows and a test set containing 275,000 flows.

As for the CTU-13 dataset, we follow the split presented by its authors~\cite{Garcia2014_an_empirical}: scenarios 3,4,5,7,10,11,12 and 13 for training and scenarios 1,2,6,8,9 for test. The training scenarios contain benign and malicious flows, so we extracted only benign flows for training. Additionally, using all data from each scenario will lead to an overly large training and test set. To address this, we sampled a subset of flows from each scenario, resulting in 640,000 flows for training and 522,000 flows for test.

For the UGR-16 dataset, we follow the split presented by its authors~\cite{Macia-Fernandez2018_ugr}: data from March to June are used as training data, while data from July and August are used as test data. As this dataset contains, by far, the largest amount of NetFlow data, we use the same sampling technique as with CTU-13 to create our training and test set. This yields a total of 600,000 flows as training data and 664,000 flows as test data.

\subsection{Methods for Comparisons}
We collected methods previously evaluated on the selected datasets to assess how SEQUENT compares to existing unsupervised and state-of-the-art methods. Additionally, we included various methods that serve as baselines.

\textbf{\textit{From Literature.}} The AssureMOSS dataset was recently published by Cao et al. and remains relatively unexplored in the context of evaluating other unsupervised anomaly detection methods. Cao et al. conducted a preliminary evaluation comparing their approach against Isolation Forest and Local Outlier Factor~\cite{cao_learning_state_machines22}. 

Regarding the CTU-13 dataset, Zoppi et al. evaluated various unsupervised methods for identifying network anomalies~\cite{Zoppi2021_unsupervised,Zoppi2021_meta_learning}. Specifically, the authors evaluated the efficacy of Isolation Forest and Fast Angle-Based Outlier Detection (FastABOD). Additionally, Grov et al. evaluated the efficacy of Markov Chains in detecting botnets~\cite{Grov19_towards}. Pellegrino et al. also conducted a study in which they evaluated the efficacy of (probabilistic) state machines and Bigram models (which are essentially Markov Chains) in detecting botnets~\cite{Pellegrino2017_learning}. 

For the UGR-16 dataset, Zoppi et al. evaluated the use of two different methods to detect network attacks~\cite{Zoppi2021_unsupervised, Zoppi2021_meta_learning}. However, due to their inability to handle the size of our train/test splits, we dropped them for comparison. Nguyen et al. utilized this dataset to evaluate the efficacy of GEE. Similar to Kitsune~\cite{Mirsky_2018_kitsune}, they use a state-of-the-art deep learning technique to detect anomalies. However, GEE explicitly trains a single VAE instead of an ensemble of AE.

We select the following from the compiled list of methods to compare against SEQUENT: Isolation Forest, FastABOD, Markov Chain, (probabilistic) state machines, and GEE. As for the implementation of the selected methods, we use Isolation Forest from the scikit-learn library~\cite{if_sklearn}, FastABOD from the PyOD library~\cite{pyod_library}, Flexfringe to learn a Markov Chain and (probabilistic) state machine, and the implementation of GEE written by Wong Mun Hou~\cite{GEE_wongmunhou}. For each method, except for the Markov Chain, we use the same parameters reported by their authors to train a model on the corresponding dataset. In cases where no parameter information is provided, we make our best effort to train a model by following the reported training steps. We chose to learn a Markov Chain using FlexFringe as we employ a distinct set of features, and learning a Markov Chain follows a standard procedure.

\textbf{\textit{Additional Methods.}}
We included additional methods beyond those previously mentioned to compare against SEQUENT. One of these methods is the straightforward Boxplot Method presented by Arp et al. ~\cite{Arp_2022_Dos}, which employs the frequency of data to detect anomalies. Furthermore, it enables us to determine whether a simple method is already sufficient to detect anomalies in our constructed test sets or if a more complex approach is required. For our experiment, we wrote our implementation of the Boxplot Method following the description provided by the authors. Zoppi et al. also used Histogram-Based Outlier Detection (HBOD) to detect network anomalies. Although not applied to any of the selected datasets, it utilizes the frequency of bins to detect anomalies. This serves as a suitable alternative baseline in our comparison. We use PyOD's implementation of HBOD. Finally, we included Kitsune as it is a state-of-the-art unsupervised ADS. We use the implementation provided by the authors~\cite{Kitsune_repo}. As it was designed to run at the packet level, we have to make minimal changes for it to work at the flow level. We recognize that this might negatively impact its performance, but we believe that it is beneficial to have this as a baseline for assessing SEQUENT's usefulness.

\subsection{Feature Selection \& FlexFringe Setup}
As selected datasets do not report the same flow features, we are required to select a subset of features shared among the selected dataset. The features shared between the selected datasets are timestamp, duration, protocol, number of bytes sent, number of packets sent, source and destination IP addresses, and source and destination ports. From this subset of features, we select the duration, protocol, number of bytes sent, and the number of packets sent to generate traces for SEQUENT. Port numbers are not considered within SEQUENT as they are spurious features within the CTU-13 dataset~\cite{nadeem_2023_sok}. We discretize all the selected features except for protocol, as it is already a categorical feature.

We configure FlexFringe to sort the flows based on the source and destination IP pairs as we want to learn a state machine from connection data. Furthermore, we configure FlexFringe to use a sliding window of length ten to generate traces for the training and test sets. By default, FlexFringe stops replaying a trace if it reaches a state and no outgoing transition exists for the next event in the trace. We, therefore, configure FlexFringe to allow the replay to reset back to the root state and continue with the remaining events. This enables us to compute the visit frequencies of the states for the complete trace. Our FlexFringe configuration settings are available on our repository~\cite{Cao_SEQUENT}.

%% file: sec_results.tex
\section{Evaluation Results}\label{sec:eval_results}


\begin{figure}[h]
    \centering
    \includegraphics[width=\columnwidth]{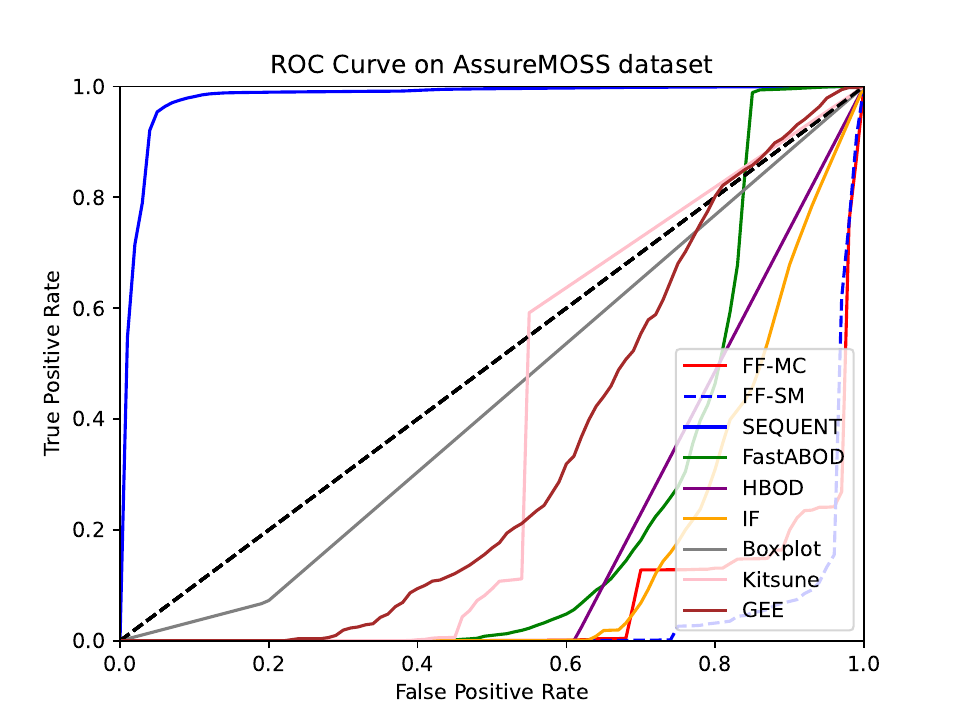}
        \caption{Average ROC curves computed for the AssureMOSS dataset.}
        \label{fig:assuremoss_roc}
\end{figure}

\begin{figure}[h]
    \centering
    \includegraphics[width=\columnwidth]{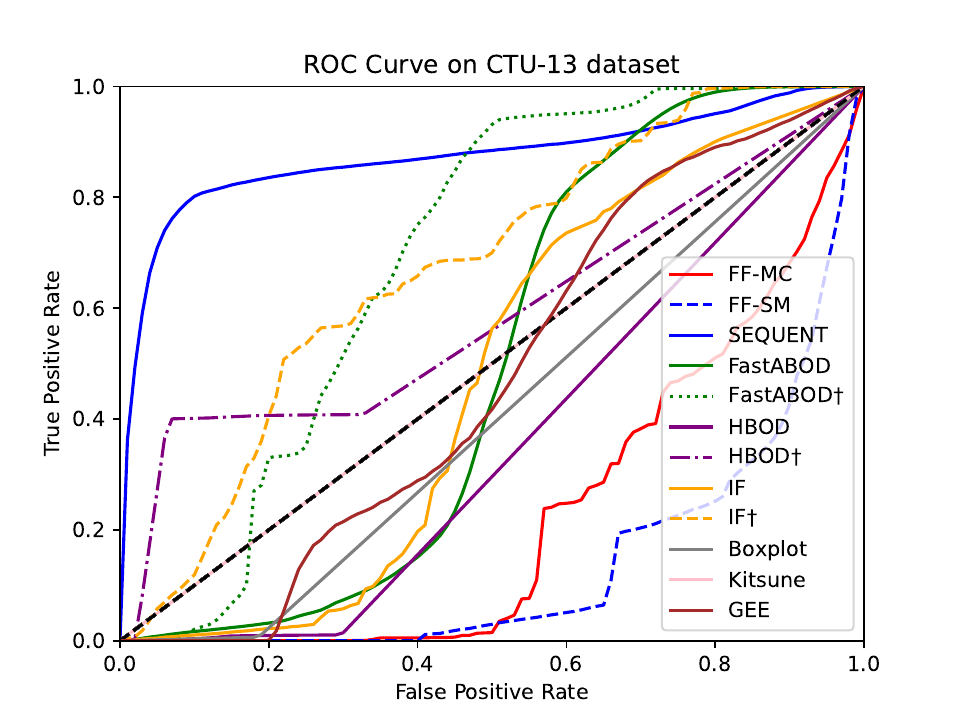}
        \caption{Average ROC curves computed for the CTU-13 dataset.}
        \label{fig:ctu_roc}
\end{figure}

\begin{figure}[h]
    \centering
    \includegraphics[width=\columnwidth]{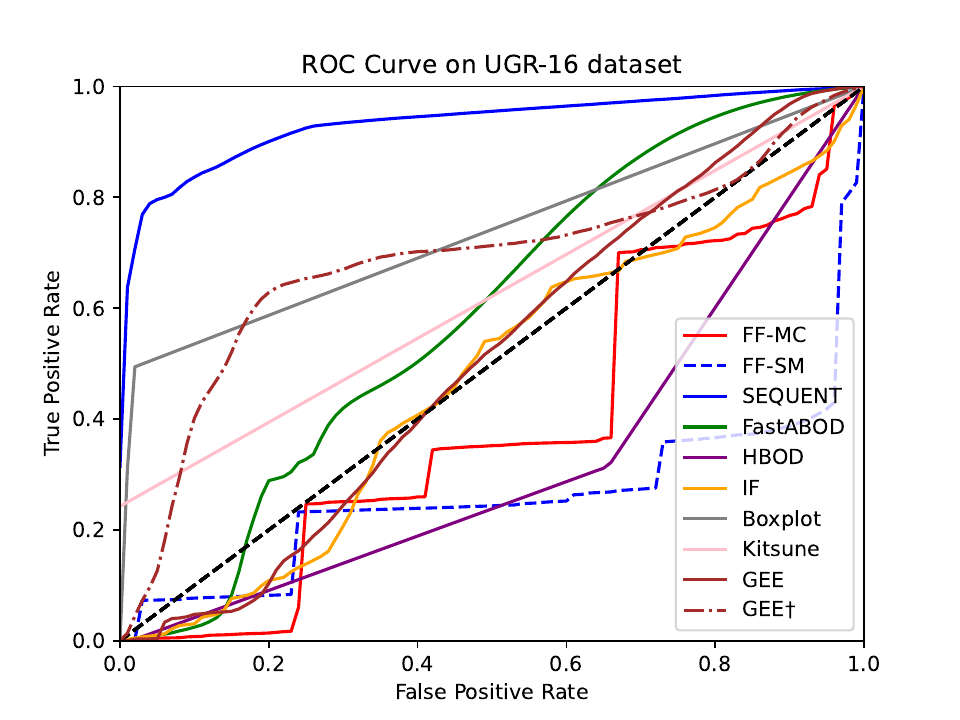}
        \caption{Average ROC curves computed for the UGR-16 dataset.}
        \label{fig:ugr_roc}
\end{figure}

\subsection{Effectiveness of SEQUENT}\label{subsec:SEQUENT_effectiveness}
We report SEQUENT's efficacy in detecting various network anomalies on the selected datasets. We use the Area Under Curve (AUC) score as the comparison metric between the various types of models. We deem this a suitable metric for comparison as each method generates anomaly scores for different traces, and we are uncertain about the optimal classification threshold that would yield the most favorable outcomes. Besides the AUC scores, we also report the Receiver Operator Characteristics (ROC) curves computed for each model. The AUC scores and ROC curves for
each dataset are reported in Table~\ref{tab:average_auc_all_datasets}, Figure~\ref{fig:assuremoss_roc}, Figure~\ref{fig:ctu_roc}, and  Figure~\ref{fig:ugr_roc}, respectively. These are average performance results computed from 10 independent runs. All models were trained on the same features considered by SEQUENT. We also include models trained on the original feature set used in the original study. These are marked with the $\dagger$-symbol in the table. Furthermore, \textit{FF-MC} denotes the Markov Chain learned using the FlexFringe framework and \textit{FF-SM} the (probabilistic) state machine. The reported results demonstrate that SEQUENT consistently outperforms all other models in detecting network anomalies on each respective dataset. Moreover, the results also suggest that SEQUENT is generalizable over different network architectures.

\begin{table}
\centering
    \begin{tabular}{l|c|c|c}
        \toprule
        \textbf{Model}             & \textbf{AssureMOSS}        & \textbf{CTU-13}       & \textbf{UGR-16 }         \\ \midrule
        Isolation Forest           & 0.140                      & 0.473                 & 0.426           \\
        Isolation Forest$^\dagger$ & -                          & 0.672                 & -               \\
        FastABOD                   & 0.229                      & 0.492                 & 0.576           \\
        FastABOD$^\dagger$         & -                          & 0.688                 & -               \\
        HBOD                       & 0.195                      & 0.357                 & 0.326           \\
        HBOD$^\dagger$             & -                          & 0.589                 & -               \\
        Boxplot Method             & 0.436                      & 0.410                 & 0.738           \\
        FF-MC                      & 0.067                      & 0.232                 & 0.392           \\
        GEE                        & 0.335                      & 0.473                 & 0.505           \\
        GEE$^\dagger$              & -                          & -                     & 0.683           \\
        Kitsune                    & 0.371                      & 0.5                   & 0.621           \\
        FF-SM                      & 0.041                      & 0.142                 & 0.257 \\
        SEQUENT                    & \textbf{0.978}             & \textbf{0.874}        & \textbf{0.933}  \\ \bottomrule
    \end{tabular}
    \caption{Average AUC of each model across all datasets.}
    \label{tab:average_auc_all_datasets}
\end{table}

\subsection{Runtime Analysis of SEQUENT}
To evaluate SEQUENT's runtime costs, we measured its training times and compared them to those of the other ML models. Additionally, we computed SEQUENT's average runtime for computing rolling anomaly scores at test time. Table~\ref{tab:sequent_training_time} presents the average training times of all ML models across all datasets. Table~\ref{tab:sequent_runtime_score} presents the average time SEQUENT takes to compute a rolling anomaly score for a given state at test time. The runtime averages were obtained from the ten independent runs of our experiments. 

From the training times presented in Table~\ref{tab:sequent_training_time}, we observe that SEQUENT takes approximately half a minute to learn a model from benign flows. While it is not the fastest of all models, it is considerably faster than models such as GEE and FastABOD. In our experiments, we used a GPU to train GEE, whereas SEQUENT does not require a GPU to learn a state machine. Although many other models have considerably lower training times, they are not as effective as SEQUENT in detecting network anomalies. This reflects a trade-off between training time and a model's detection performance. We believe that spending a bit more time to train SEQUENT can be justified, as it achieves a considerably higher detection performance than the other models. We further support our belief based on SEQUENT's runtimes in computing rolling anomalies scores across the datasets (see Table~\ref{tab:sequent_runtime_score}. SEQUENT computes anomaly scores in microseconds on average, making it a practical solution for network anomaly detection.

\begin{table}
\centering
    \begin{tabular}{l|c|c|c}
        \toprule
        \textbf{Model}             & \textbf{AssureMOSS}        & \textbf{CTU-13}       & \textbf{UGR-16 }      \\ \midrule
        Isolation Forest           & 0.18  $\pm~0.01$            & 1.34   $\pm~0.11$     & 1.00    $\pm~0.04$    \\
        Isolation Forest$^\dagger$ & -                          & 1.26  $\pm~0.13$      & -                     \\
        FastABOD                   & 16.13 $\pm~0.65$           & 88.30 $\pm~1.26$      & 82.34   $\pm~2.83$    \\
        FastABOD$^\dagger$         & -                          & 101.29  $\pm~6.92$     & -                     \\
        HBOD                       & 0.02  $\pm~0.0$            & 0.25   $\pm~0.50$     & 0.10    $\pm~0.0$     \\
        HBOD$^\dagger$             & -                          & 0.27   $\pm~0.54$     & -                     \\
        Boxplot Method             & 7.24  $\pm~0.25$           & 67.14  $\pm~0.41$     & 100.50  $\pm~0.53$    \\
        FF-MC                      & 1.07  $\pm~0.03$           & 6.51   $\pm~0.13$     & 5.22    $\pm~0.09$    \\
        GEE                        & 99.00 $\pm~0.10$           & 127.44 $\pm~0.57$     & 169.69  $\pm~0.57$    \\
        GEE$^\dagger$              & -                          & -                     & 169.55  $\pm~0.35$    \\
        Kitsune                    & 1.24  $\pm~0.08$           & 6.38  $\pm~0.26$      & 6.06  $\pm~0.22$      \\
        FF-SM                      & 4.02  $\pm~0.12$           & 17.41 $\pm~1.18$      & 22.88   $\pm~1.53$    \\
        SEQUENT                    & 5.11  $\pm~0.30$           & 19.23 $\pm~0.22$      & 27.21   $\pm~2.55$    \\ \bottomrule
    \end{tabular}
    \caption{Average training time (seconds) computed across all datasets.}
    \label{tab:sequent_training_time}
\end{table}

\begin{table}
\centering
\resizebox{\columnwidth}{!}{
    \begin{tabular}{@{}l|c@{}}
    \toprule
    \textbf{Dataset} & \textbf{Avg. Runtime for Score Computation} \\ \midrule
    AssureMOSS       & 2.17 $\pm~3.96$                                    \\
    CTU-13           & 2.20 $\pm~5.17$                                    \\
    UGR-16           & 2.36 $\pm~3.62$                                    \\ \bottomrule
    \end{tabular}
}
\caption{Average runtime (microseconds) for computing rolling anomaly score for a state. }
\label{tab:sequent_runtime_score}
\end{table}

\begin{figure*}[t!]
    \centering
    \includegraphics[width=\textwidth]{figures/ctu_simplified_model.pdf}
    \caption{Simplified state machine model learned from CTU-13 dataset.}
    \label{fig:state_machine_ctu}
\end{figure*}

\begin{figure}[t!]
    \centering
    \includegraphics[width=\columnwidth]{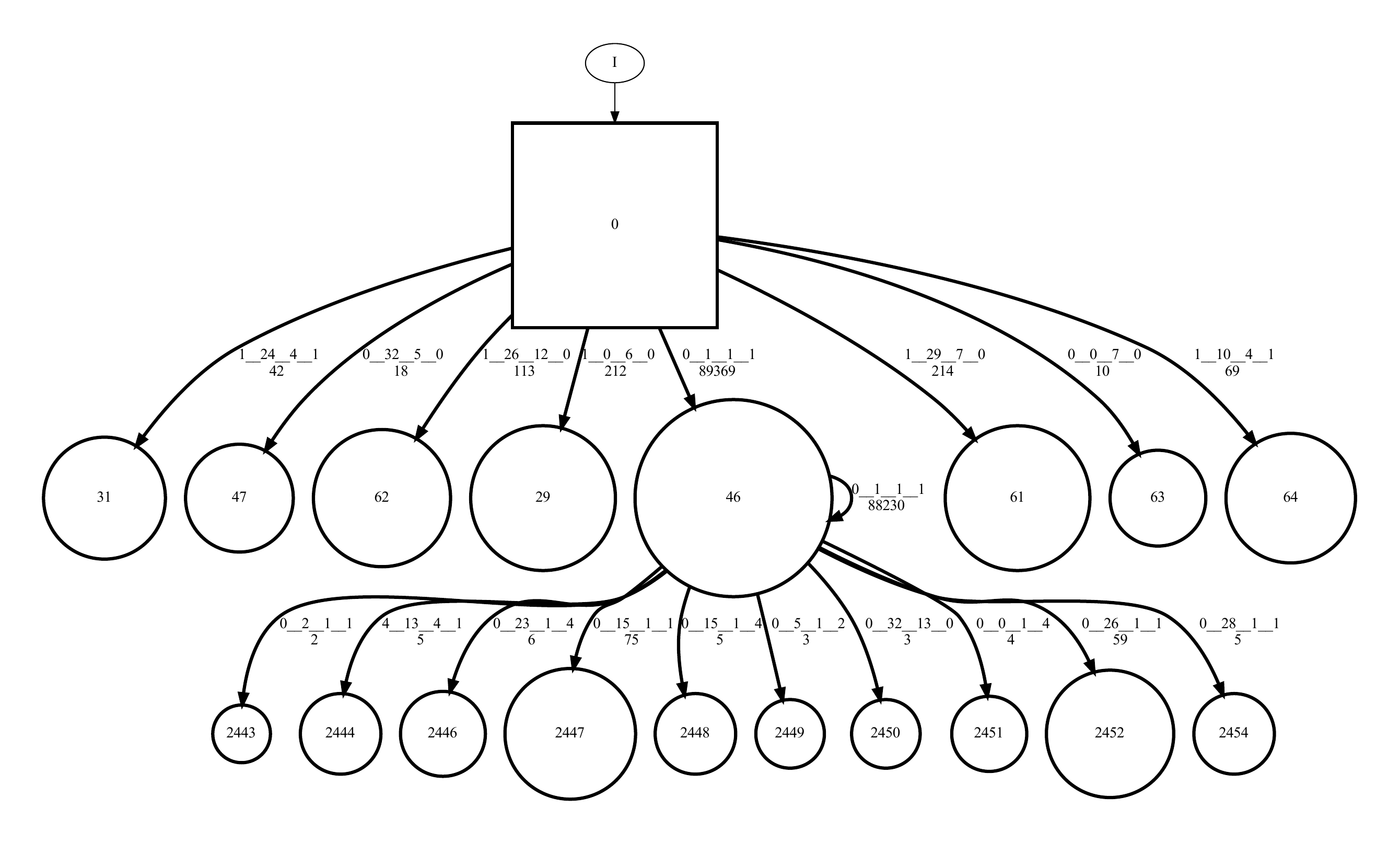}
    \caption{Zoomed-in state machine of Figure~\ref{fig:state_machine_ctu}, showing the top three anomalous states (47, 62 and 31) presented in Table~\ref{table:ctu_rc_stats}.}
    \label{fig:state_machine_ctu_zoomed_in}
\end{figure}

\subsection{Example Analysis Using Root Causes}
One of SEQUENT's advantages is the ability to group anomalies using the generated root-cause symbols. This enables an analyst to reduce the time spent analyzing false positives by filtering out symbols deemed to be false positives. We provide an illustrative example to demonstrate how these symbols can aid with the analysis of anomalies. Let's say that an analyst has set an arbitrary threshold on the (rolling) anomaly scores to select \textit{"interesting"} alerts for further analysis. The analyst then groups the selected traces based on their root-cause symbols and computes the size of each group. Table~\ref{table:ctu_rc_stats} presents the top 3 largest groups of anomalies computed from the CTU-13 dataset. Additionally, Figure~\ref{fig:state_machine_ctu} illustrates the simplified state machine learned by SEQUENT. The analyst decides to first analyze group "\textit{62}" by investigating the corresponding (starting) flows collected and linked by SEQUENT (see Table~\ref{table:ctu_rc_62}). The most anomalous flows are shown at the top. SEQUENT deems these flows to be anomalous due to the significant increase in network behavior that coincidentally correlates with the use of port 25, which is used for Simple Mail Transfer Protocol (SMTP). The analyst deems these flows to be suspicious due to the destination port used in the communication and marks all traces with the root-cause symbol "\textit{62}" as malicious. Examining the actual labels of these flows, they correspond to the malicious traffic produced by the Neris botnet. This malware specifically sends spam mail~\cite{Garcia2014_an_empirical}, which explains the use of the SMTP port (25).

\begin{table}
\centering
    \begin{tabular}{c|c}
    \toprule
    \textbf{Root-Cause Symbol} & \textbf{Size} \\ \midrule
    62                                              & 18646          \\
    47                                              & 3182           \\
    31                                              & 2556           \\ 
    1501                                            & 2448           \\
    72                                              & 2244           \\ \bottomrule
    \end{tabular}
\caption{Anomaly group sizes computed using root-cause symbols generated for the CTU-13 dataset}
\label{table:ctu_rc_stats}
\end{table}


\begin{table}
\centering
\resizebox{\columnwidth}{!}{
\begin{tabular}{c|c|c|c|c|c}
\toprule
\textbf{src\_ip}    & \textbf{dst\_ip} & \textbf{sport} & \textbf{dport}    & \textbf{num\_bytes}   & \textbf{num\_packets} \\ \midrule
147.32.84.165       & 66.94.237.64     & 3271  & 25     & 186               & 3                     \\
147.32.84.165       & 98.139.175.225   & 3539  & 25     & 186               & 3                     \\
147.32.84.165       & 205.188.146.194  & 2474  & 25     & 186               & 3                     \\
147.32.84.165       & 76.96.30.116     & 3268  & 25     & 186               & 3                     \\
147.32.84.165       & 216.157.130.15   & 1490  & 25     & 186               & 3                     \\
147.32.84.165       & 98.137.54.237    & 3630  & 25     & 186               & 3                     \\
147.32.84.165       & 64.12.138.161    & 1789  & 25     & 186               & 3                     \\
147.32.84.165       & 209.85.143.27    & 3328  & 25     & 186               & 3                     \\
147.32.84.165       & 205.188.190.2    & 3857  & 25     & 186               & 3                     \\
147.32.84.165       & 98.139.175.225   & 3508  & 25     & 186               & 3                     \\ \bottomrule
\end{tabular}}
\caption{Simplified flows linked to the top 10 anomalous traces for root-cause symbol "62".} 
\label{table:ctu_rc_62}
\end{table}

Following the same approach, the analyst now investigates group "\textit{47}" using the linked flows depicted in Table~\ref{table:ctu_rc_47}. SEQUENT does not know the actual nature behind these flows, but due to the significant increase in this type of network behavior, it deems these flows to be anomalous. Upon inspection, the analyst observes that these are correlated with the use of port 53, which is typically used for domain name resolution (DNS). This significant increase could be related to a DNS flood attack, or it could simply be that this type of network behavior was rarely seen during training. The analyst deems these flows unrelated to an attack and marks all traces grouped under the symbol "47" as false positives. By investigating only a subset of the linked flows, the analyst can filter out 3182 false positive alerts and reduce the time spent analyzing every alert. Furthermore, the analyst can visualize the part of the model causing the most anomalous activity (see Figure~\ref{fig:state_machine_ctu_zoomed_in}) to understand the sequential behavior exhibited by the most anomalous groups of traces. 

\begin{table}
\centering
\resizebox{\columnwidth}{!}{
\begin{tabular}{c|c|c|c|c|c}
\toprule
\textbf{src\_ip} & \textbf{dst\_ip} & \textbf{sport} & \textbf{dport} & \textbf{num\_bytes} & \textbf{num\_packets} \\ \midrule
147.32.84.138    & 147.32.80.9      & 45170          & 53             & 214                 & 2                    \\
147.32.84.138    & 147.32.80.9      & 57876          & 53             & 214                 & 2                    \\
147.32.84.138    & 147.32.80.9      & 50385          & 53             & 214                 & 2                    \\
147.32.84.138    & 147.32.80.9      & 45753          & 53             & 214                 & 2                    \\
147.32.84.138    & 147.32.80.9      & 34683          & 53             & 214                 & 2                    \\
147.32.84.138    & 147.32.80.9      & 42564          & 53             & 214                 & 2                    \\
147.32.84.138    & 147.32.80.9      & 56081          & 53             & 214                 & 2                    \\
147.32.84.138    & 147.32.80.9      & 58016          & 53             & 214                 & 2                    \\
147.32.84.138    & 147.32.80.9      & 53136          & 53             & 214                 & 2                    \\
147.32.84.138    & 147.32.80.9      & 40272          & 53             & 214                 & 2                    \\ \bottomrule
\end{tabular}}
\caption{Simplified flows linked to the top 10 anomalous traces for root-cause symbol "47".} 
\label{table:ctu_rc_47}
\end{table}

\subsection{SEQUENT's Robustness}
An essential characteristic of an effective ADS is its ability to withstand attacks launched by adversaries seeking to evade detection. With various works showing that ML models are susceptible to adversarial attacks~\cite{apruzzese_2019_addressing,biggio_2013_evasion,szegedy2014intriguing}, it is imperative to evaluate SEQUENT's efficacy against (potential) adversarial attacks. To so do, we constructed four distinct types of adversarial attacks following the adversary's capabilities presented in Section~\ref{sec:adversary_assumptions}. We strictly consider attacks that an adversary could launch at test time, and we do not consider adversarial attacks such as poisoning of training data. An adversary is only allowed to adapt the features considered by SEQUENT for anomaly detection, and they are allowed to collect benign flows from the targeted network before the start of an attack. The constructed adversarial attacks are defined as follows:

\begin{itemize}
    \item \textbf{\textit{Padding Attack}}: The adversary is allowed to mimic normal behavior by adapting each feature value to its nearest value within the collected data. 
    \item \textbf{\textit{Random Replacement Attack}}: The adversary mimics normal behavior by replacing their flows with random flows selected from the collected data.
    \item \textbf{\textit{Window Replacement Attack}}: The adversary mimics normal behavior by replacing each sliding window (10 flows) of their flows with a randomly selected window of benign flows. Each window is selected at most once.  
    \item \textbf{\textit{Frequency Replacement Attack}}: The adversary mimics normal behavior by finding frequently occurring flows (at least 100 times) from the collected data and replaces their flows with randomly selected frequently occurring flows. The intuition behind this attack is that an adversary would be able to blend in well with the normal flows if they mimic the frequently occurring communication behavior. 
\end{itemize}

Tables~\ref{table:avg_auc_all_models_robust} presents SEQUENT's robustness against the aforementioned adversarial attacks on each selected dataset. The results are the average AUC scores across 10 independent runs. 
Based on the reported results, we see that SEQUENT achieved the best performance against two types of adversarial attacks (\textit{Padding Attack} and \textit{Random Replacement Attack}) compared to all other models. This can be explained by the nature of these two attacks; when an adversary tries to mimic the normal network behavior by padding or doing replacements on their flows, the traces will resemble the traces observed at training. However, this also changes the sequential behavior exhibited in the traces, which also results in changes in the state visit frequencies during the replay. SEQUENT detects this unusual increase in visit frequencies in the states and will raise an alarm. Existing methods trained on the raw feature value will not detect this deviation as these values are similar to what was observed during training. Probabilistic models such as Markov Chains and probabilistic state machines also suffer from this issue as the adversarial traces are similar to what was observed during training. The models will assign higher probabilities to these traces, thus resulting in poor detection performance.


Another interesting observation is that SEQUENT's performance deteriorates significantly on the AssureMOSS and UGR-16 datasets but not on the CTU-13 dataset when an adversary launches a \textit{Window Replacement Attack}. This decline in performance is expected since all malicious windows are randomly replaced by benign windows. Unlike the other three attacks, flows or feature values are not randomly replaced; instead, the flows and their sequential behaviors resemble the ones seen during training. As each window is allowed to be selected at most once, the sequential behavior exhibited in the (copied) window occurs at a slightly higher frequency, though not significantly more than observed during training. The complexity behind this attack makes it challenging for SEQUENT to detect the adversary. However, SEQUENT exhibits robustness against this specific attack on the CTU-13 dataset. We speculate that this result is influenced by the type of attack captured in the dataset. 


One discussion point that we would like to initiate concerning three of the constructed adversarial attacks (in which replacement is involved) is whether an adversary could still accomplish their objectives by producing exact copies of the normal traffic data to avoid detection by an ADS. We argue that by producing literal copies and even going to the extent of recreating the sequential behaviors of benign flows, the adversary is essentially replaying collected network traffic data. No system, not even the analyst doing manual inspection of each flow or packet, would be able to explain how the data produced by an adversary differs from the normal traffic data as it is a literal copy. 

\begin{table}
\resizebox{\columnwidth}{!}{
\begin{tabular}{l|c|c|c|c}
\toprule
\textbf{Model}              & \textbf{Padding}        & \textbf{Random}         & \textbf{Window}         & \textbf{Frequency}      \\ \midrule
\multicolumn{5}{c}{\textbf{AssureMOSS}}                                                                                                                                                            \\ \midrule
Isolation Forest            & 0.162          & 0.185          & 0.431          & 0.155          \\
FastABOD                    & 0.222          & 0.220          & 0.177          & 0.031          \\
HBOD                        & 0.195          & 0.226          & 0.404          & 0.195          \\
GEE                         & 0.576          & 0.441          & 0.427          & 0.916          \\
Kitsune                     & 0.859          & 0.862          & \textbf{0.573} & 0.835          \\
FF-MC                       & 0.014          & 0.062          & 0.0319         & \textbf{0.985} \\
FF-SM                       & 0.007          & 0.038          & 0.278          & \textbf{0.985} \\
Boxplot                     & 0.435          & 0.465          & 0.502          & 0.441          \\
SEQUENT                     & \textbf{0.991} & \textbf{0.903} & 0.155          & 0.771          \\ \midrule
\multicolumn{5}{c}{\textbf{CTU-13}}                                                                                                                                                                \\ \midrule
Isolation Forest & 0464                               & 0.484                              & 0.515                              & 0.478                              \\
FastABOD         & 0.492                              & 0.475                              & 0.436                              & 0.290                              \\
HBOD             & 0.357                              & 0.369                              & 0.392                              & 0.497                              \\
GEE              & 0.470                              & 0.372                              & 0.522                              & 0.490                              \\
Kitsune          & 0.5                                & 0.5                                & 0.5                                & 0.5                                \\
FF-MC            & 0.231                              & 0.2                                & 0.283                              & 0.298                              \\
FF-SM            & 0.142                              & 0.091                              & 0.323                              & 0.441                              \\
Boxplot          & 0.5                                & 0.5                                & 0.5                                & 0.5                                \\
SEQUENT          & \textbf{0.970}                     & \textbf{0.966}                     & \textbf{0.861}                     & \textbf{0.974}                     \\ \midrule
\multicolumn{5}{c}{\textbf{UGR-16}}                                                                                                                                                                \\ \midrule
Isolation Forest & 0.431                              & 0.429                              & 0.480                              & 0.470                              \\
FastABOD         & 0.588                              & 0.566                              & 0.468                              & 0.336                              \\
HBOD             & 0.327                              & 0.333                              & 0.364                              & 0.373                              \\
GEE              & 0.115                              & 0.122                              & 0.113                              & 0.052                              \\
Kitsune          & 0.621                              & 0.621                              & \textbf{0.621}                     & 0.621                              \\
FF-MC            & 0.323                              & 0.308                              & 0.491                              & 0.640                              \\
FF-SM            & 0.174                              & 0.166                              & 0.518                              & 0.811                              \\
Boxplot          & 0.740                              & 0.758                              & 0.426                              & 0.742                              \\
SEQUENT          & \textbf{0.983}                     & \textbf{0.919}                     & 0.407                              & \textbf{0.927}                     \\ \bottomrule
\end{tabular}
}
\caption{The performance (average AUC) of each model against the four adversarial attacks.}
\label{table:avg_auc_all_models_robust}
\end{table}

%% file: sec_conclusion_future.tex
\section{Conclusion \& Future Work}\label{sec:conclusion_future}
We introduced SEQUENT, a framework that employs state machines to detect various types of network anomalies. SEQUENT leverages the structure of a state machine to collect state visit frequencies and subsequently uses them to continually adapt its scoring for arbitrary traces seen at test time. We observed that SEQUENT achieves a better overall performance in detecting various types of network anomalies, even outperforming existing state-of-the-art unsupervised anomaly detection methods. 
Furthermore, we demonstrated how SEQUENT computes root causes for the detected anomalies and uses them to group and rank anomalies. Moreover, SEQUENT links root causes back to the concrete NetFlow instances. The root causes, and the linked instances enable an analyst to quickly filter out a large number of false alerts, reducing the time spent on analyzing them. Finally, we evaluated SEQUENT's robustness against four types of adversarial attacks. Reported results show that SEQUENT achieved the best performance in two of the adversarial attacks.

One potential future research direction is to investigate how the intuition underlying SEQUENT's scoring method can be applied to other existing models, such as isolation forests or neural networks. It is not trivial how the internal structure of these models can be employed to continuously adapt its anomaly scores for new data seen at test time. 

Another research direction is to apply SEQUENT to software log data. We believe that SEQUENT might also be effective in detecting anomalies in software logs, and it would be interesting to compare SEQUENT against existing anomaly detection methods developed for software log data.

%% file: acm_main.bbl

\begin{thebibliography}{36}


\ifx \showCODEN    \undefined \def \showCODEN     #1{\unskip}     \fi
\ifx \showDOI      \undefined \def \showDOI       #1{#1}\fi
\ifx \showISBNx    \undefined \def \showISBNx     #1{\unskip}     \fi
\ifx \showISBNxiii \undefined \def \showISBNxiii  #1{\unskip}     \fi
\ifx \showISSN     \undefined \def \showISSN      #1{\unskip}     \fi
\ifx \showLCCN     \undefined \def \showLCCN      #1{\unskip}     \fi
\ifx \shownote     \undefined \def \shownote      #1{#1}          \fi
\ifx \showarticletitle \undefined \def \showarticletitle #1{#1}   \fi
\ifx \showURL      \undefined \def \showURL       {\relax}        \fi
\providecommand\bibfield[2]{#2}
\providecommand\bibinfo[2]{#2}
\providecommand\natexlab[1]{#1}
\providecommand\showeprint[2][]{arXiv:#2}

\bibitem[Alahmadi et~al\mbox{.}(2022)]%
        {bushra2022_99false}
\bibfield{author}{\bibinfo{person}{Bushra~A. Alahmadi}, \bibinfo{person}{Louise Axon}, {and} \bibinfo{person}{Ivan Martinovic}.} \bibinfo{year}{2022}\natexlab{}.
\newblock \showarticletitle{99\% False Positives: A Qualitative Study of {SOC} Analysts{\textquoteright} Perspectives on Security Alarms}. In \bibinfo{booktitle}{\emph{31st USENIX Security Symposium}}. \bibinfo{publisher}{USENIX Association}, \bibinfo{address}{Boston, MA}, \bibinfo{pages}{2783--2800}.
\newblock
\showISBNx{978-1-939133-31-1}
\urldef\tempurl%
\url{https://www.usenix.org/conference/usenixsecurity22/presentation/alahmadi}
\showURL{%
\tempurl}


\bibitem[Angluin(1987)]%
        {Angluin_1987_learning}
\bibfield{author}{\bibinfo{person}{Dana Angluin}.} \bibinfo{year}{1987}\natexlab{}.
\newblock \showarticletitle{Learning regular sets from queries and counterexamples}.
\newblock \bibinfo{journal}{\emph{Information and Computation}} \bibinfo{volume}{75}, \bibinfo{number}{2} (\bibinfo{year}{1987}), \bibinfo{pages}{87--106}.
\newblock
\showISSN{0890-5401}
\urldef\tempurl%
\url{https://doi.org/10.1016/0890-5401(87)90052-6}
\showDOI{\tempurl}


\bibitem[Apruzzese et~al\mbox{.}(2019)]%
        {apruzzese_2019_addressing}
\bibfield{author}{\bibinfo{person}{Giovanni Apruzzese}, \bibinfo{person}{Michele Colajanni}, \bibinfo{person}{Luca Ferretti}, {and} \bibinfo{person}{Mirco Marchetti}.} \bibinfo{year}{2019}\natexlab{}.
\newblock \showarticletitle{Addressing Adversarial Attacks Against Security Systems Based on Machine Learning}. In \bibinfo{booktitle}{\emph{2019 11th International Conference on Cyber Conflict (CyCon)}}, Vol.~\bibinfo{volume}{900}. \bibinfo{pages}{1--18}.
\newblock
\urldef\tempurl%
\url{https://doi.org/10.23919/CYCON.2019.8756865}
\showDOI{\tempurl}


\bibitem[Arp et~al\mbox{.}(2022)]%
        {Arp_2022_Dos}
\bibfield{author}{\bibinfo{person}{Daniel Arp}, \bibinfo{person}{Erwin Quiring}, \bibinfo{person}{Feargus Pendlebury}, \bibinfo{person}{Alexander Warnecke}, \bibinfo{person}{Fabio Pierazzi}, \bibinfo{person}{Christian Wressnegger}, \bibinfo{person}{Lorenzo Cavallaro}, {and} \bibinfo{person}{Konrad Rieck}.} \bibinfo{year}{2022}\natexlab{}.
\newblock \showarticletitle{Dos and don'ts of machine learning in computer security}. In \bibinfo{booktitle}{\emph{31st USENIX Security Symposium}}. \bibinfo{pages}{3971--3988}.
\newblock


\bibitem[Biggio et~al\mbox{.}(2013)]%
        {biggio_2013_evasion}
\bibfield{author}{\bibinfo{person}{Battista Biggio}, \bibinfo{person}{Igino Corona}, \bibinfo{person}{Davide Maiorca}, \bibinfo{person}{Blaine Nelson}, \bibinfo{person}{Nedim {\v{S}}rndi{\'{c}}}, \bibinfo{person}{Pavel Laskov}, \bibinfo{person}{Giorgio Giacinto}, {and} \bibinfo{person}{Fabio Roli}.} \bibinfo{year}{2013}\natexlab{}.
\newblock \showarticletitle{Evasion Attacks against Machine Learning at Test Time}. In \bibinfo{booktitle}{\emph{Machine Learning and Knowledge Discovery in Databases}}, \bibfield{editor}{\bibinfo{person}{Hendrik Blockeel}, \bibinfo{person}{Kristian Kersting}, \bibinfo{person}{Siegfried Nijssen}, {and} \bibinfo{person}{Filip {\v{Z}}elezn{\'y}}} (Eds.). \bibinfo{publisher}{Springer Berlin Heidelberg}, \bibinfo{address}{Berlin, Heidelberg}, \bibinfo{pages}{387--402}.
\newblock
\showISBNx{978-3-642-40994-3}


\bibitem[Buitinck et~al\mbox{.}(2024)]%
        {if_sklearn}
\bibfield{author}{\bibinfo{person}{Lars Buitinck}, \bibinfo{person}{Gilles Louppe}, \bibinfo{person}{Mathieu Blondel}, \bibinfo{person}{Fabian Pedregosa}, \bibinfo{person}{Andreas Mueller}, \bibinfo{person}{Olivier Grisel}, \bibinfo{person}{Vlad Niculae}, \bibinfo{person}{Peter Prettenhofer}, \bibinfo{person}{Alexandre Gramfort}, \bibinfo{person}{Jaques Grobler}, \bibinfo{person}{Robert Layton}, \bibinfo{person}{Jake VanderPlas}, \bibinfo{person}{Arnaud Joly}, \bibinfo{person}{Brian Holt}, {and} \bibinfo{person}{Ga{\"{e}}l Varoquaux}.} \bibinfo{year}{2024}\natexlab{}.
\newblock \bibinfo{title}{{sklearn.ensemble.IsolationForest — scikit-learn 1.1.2 documentation}}.
\newblock \bibinfo{howpublished}{\url{https://scikit-learn.org/stable/modules/generated/sklearn.ensemble.IsolationForest.html}}.
\newblock


\bibitem[Campazas-Vega et~al\mbox{.}(2023)]%
        {Campazas_2023_Malicious}
\bibfield{author}{\bibinfo{person}{Adri{\'a}n Campazas-Vega}, \bibinfo{person}{Ignacio~Samuel Crespo-Mart{\'\i}nez}, \bibinfo{person}{{\'A}ngel~Manuel Guerrero-Higueras}, \bibinfo{person}{Claudia {\'A}lvarez-Aparicio}, \bibinfo{person}{Vicente Matell{\'a}n}, {and} \bibinfo{person}{Camino Fern{\'a}ndez-Llamas}.} \bibinfo{year}{2023}\natexlab{}.
\newblock \showarticletitle{Malicious traffic detection on sampled network flow data with novelty-detection-based models}.
\newblock \bibinfo{journal}{\emph{Scientific Reports}} \bibinfo{volume}{13}, \bibinfo{number}{1} (\bibinfo{year}{2023}), \bibinfo{pages}{15446}.
\newblock


\bibitem[Cao(2022a)]%
        {encode_repo}
\bibfield{author}{\bibinfo{person}{Clinton Cao}.} \bibinfo{year}{2022}\natexlab{a}.
\newblock \bibinfo{title}{{ENCODE: Encoding NetFlows for Network Anomaly Detection}}.
\newblock \bibinfo{howpublished}{\url{https://github.com/tudelft-cda-lab/ENCODE}}.
\newblock


\bibitem[Cao(2022b)]%
        {Cao_SEQUENT}
\bibfield{author}{\bibinfo{person}{Clinton Cao}.} \bibinfo{year}{2022}\natexlab{b}.
\newblock \bibinfo{booktitle}{\emph{{SEQEUENT: State Frequency Estimation for Anomaly Detection}}}.
\newblock


\bibitem[Cao and Blaise(2022)]%
        {assuremoss_dataset}
\bibfield{author}{\bibinfo{person}{Clinton Cao} {and} \bibinfo{person}{Agathe Blaise}.} \bibinfo{year}{2022}\natexlab{}.
\newblock \bibinfo{title}{{AssureMOSS Kubernetes Run-time Monitoring Dataset}}.
\newblock \bibinfo{howpublished}{\url{https://data.4tu.nl/articles/\_/20463687/1}}.
\newblock
\urldef\tempurl%
\url{https://doi.org/10.4121/20463687.V1}
\showDOI{\tempurl}


\bibitem[Cao et~al\mbox{.}(2022)]%
        {cao_learning_state_machines22}
\bibfield{author}{\bibinfo{person}{Clinton Cao}, \bibinfo{person}{Agathe Blaise}, \bibinfo{person}{Sicco Verwer}, {and} \bibinfo{person}{Filippo Rebecchi}.} \bibinfo{year}{2022}\natexlab{}.
\newblock \showarticletitle{Learning State Machines to Monitor and Detect Anomalies on a Kubernetes Cluster}. In \bibinfo{booktitle}{\emph{Proceedings of the 17th International Conference on Availability, Reliability and Security}} (Vienna, Austria) \emph{(\bibinfo{series}{ARES '22})}. \bibinfo{publisher}{Association for Computing Machinery}, \bibinfo{address}{New York, NY, USA}, Article \bibinfo{articleno}{117}, \bibinfo{numpages}{9}~pages.
\newblock
\showISBNx{9781450396707}
\urldef\tempurl%
\url{https://doi.org/10.1145/3538969.3543810}
\showDOI{\tempurl}


\bibitem[Fosić et~al\mbox{.}(2023)]%
        {Fosic_2023_Anomaly}
\bibfield{author}{\bibinfo{person}{Igor Fosić}, \bibinfo{person}{Drago Žagar}, \bibinfo{person}{Krešimir Grgić}, {and} \bibinfo{person}{Višnja Križanović}.} \bibinfo{year}{2023}\natexlab{}.
\newblock \showarticletitle{Anomaly detection in NetFlow network traffic using supervised machine learning algorithms}.
\newblock \bibinfo{journal}{\emph{Journal of Industrial Information Integration}}  \bibinfo{volume}{33} (\bibinfo{year}{2023}), \bibinfo{pages}{100466}.
\newblock
\showISSN{2452-414X}
\urldef\tempurl%
\url{https://doi.org/10.1016/j.jii.2023.100466}
\showDOI{\tempurl}


\bibitem[Garcia et~al\mbox{.}(2014)]%
        {Garcia2014_an_empirical}
\bibfield{author}{\bibinfo{person}{S. Garcia}, \bibinfo{person}{M. Grill}, \bibinfo{person}{J. Stiborek}, {and} \bibinfo{person}{A. Zunino}.} \bibinfo{year}{2014}\natexlab{}.
\newblock \showarticletitle{{An empirical comparison of botnet detection methods}}.
\newblock \bibinfo{journal}{\emph{Computers \& Security}} (\bibinfo{date}{sep} \bibinfo{year}{2014}).
\newblock
\showISSN{0167-4048}
\urldef\tempurl%
\url{https://doi.org/10.1016/J.COSE.2014.05.011}
\showDOI{\tempurl}


\bibitem[Grov et~al\mbox{.}(2019)]%
        {Grov19_towards}
\bibfield{author}{\bibinfo{person}{Gudmund Grov}, \bibinfo{person}{Marc Sabate}, \bibinfo{person}{Wei Chen}, {and} \bibinfo{person}{David Aspinall}.} \bibinfo{year}{2019}\natexlab{}.
\newblock \showarticletitle{Towards Intelligible Robust Anomaly Detection by Learning Interpretable Behavioural Models}.
\newblock \bibinfo{journal}{\emph{NISK J}}  \bibinfo{volume}{32} (\bibinfo{year}{2019}), \bibinfo{pages}{1--16}.
\newblock


\bibitem[Hassan et~al\mbox{.}(2024)]%
        {Waseem_2024_Real}
\bibfield{author}{\bibinfo{person}{Waseem Hassan}, \bibinfo{person}{Seyed~Ebrahim Hosseini}, {and} \bibinfo{person}{Shahbaz Pervez}.} \bibinfo{year}{2024}\natexlab{}.
\newblock \showarticletitle{Real-Time Anomaly Detection in Network Traffic Using Graph Neural Networks and Random Forest}. In \bibinfo{booktitle}{\emph{Internet of Things, Smart Spaces, and Next Generation Networks and Systems}}, \bibfield{editor}{\bibinfo{person}{Yevgeni Koucheryavy} {and} \bibinfo{person}{Ahmed Aziz}} (Eds.). \bibinfo{publisher}{Springer Nature Switzerland}, \bibinfo{address}{Cham}, \bibinfo{pages}{194--207}.
\newblock
\showISBNx{978-3-031-60994-7}


\bibitem[Lang et~al\mbox{.}(1998)]%
        {lang_1998_results}
\bibfield{author}{\bibinfo{person}{Kevin~J. Lang}, \bibinfo{person}{Barak~A. Pearlmutter}, {and} \bibinfo{person}{Rodney~A. Price}.} \bibinfo{year}{1998}\natexlab{}.
\newblock \showarticletitle{Results of the Abbadingo one DFA learning competition and a new evidence-driven state merging algorithm}. In \bibinfo{booktitle}{\emph{Grammatical Inference}}, \bibfield{editor}{\bibinfo{person}{Vasant Honavar} {and} \bibinfo{person}{Giora Slutzki}} (Eds.). \bibinfo{publisher}{Springer Berlin Heidelberg}, \bibinfo{address}{Berlin, Heidelberg}, \bibinfo{pages}{1--12}.
\newblock
\showISBNx{978-3-540-68707-8}


\bibitem[Lanvin et~al\mbox{.}(2023)]%
        {lanvin2023_towards}
\bibfield{author}{\bibinfo{person}{Maxime Lanvin}, \bibinfo{person}{Pierre-Fran\c{c}ois Gimenez}, \bibinfo{person}{Yufei Han}, \bibinfo{person}{Fr\'{e}d\'{e}ric Majorczyk}, \bibinfo{person}{Ludovic M\'{e}}, {and} \bibinfo{person}{Eric Totel}.} \bibinfo{year}{2023}\natexlab{}.
\newblock \showarticletitle{Towards Understanding Alerts Raised by Unsupervised Network Intrusion Detection Systems}. In \bibinfo{booktitle}{\emph{Proceedings of the 26th International Symposium on Research in Attacks, Intrusions and Defenses}} (Hong Kong, China) \emph{(\bibinfo{series}{RAID '23})}. \bibinfo{publisher}{Association for Computing Machinery}, \bibinfo{address}{New York, NY, USA}, \bibinfo{pages}{135–150}.
\newblock
\showISBNx{9798400707650}
\urldef\tempurl%
\url{https://doi.org/10.1145/3607199.3607247}
\showDOI{\tempurl}


\bibitem[Lee(2021)]%
        {Lee_2021_determinism}
\bibfield{author}{\bibinfo{person}{Edward~A. Lee}.} \bibinfo{year}{2021}\natexlab{}.
\newblock \showarticletitle{Determinism}.
\newblock \bibinfo{journal}{\emph{ACM Trans. Embed. Comput. Syst.}} \bibinfo{volume}{20}, \bibinfo{number}{5}, Article \bibinfo{articleno}{38} (\bibinfo{date}{may} \bibinfo{year}{2021}), \bibinfo{numpages}{34}~pages.
\newblock
\showISSN{1539-9087}
\urldef\tempurl%
\url{https://doi.org/10.1145/3453652}
\showDOI{\tempurl}


\bibitem[Lin et~al\mbox{.}(2018)]%
        {Lin18_TABOR}
\bibfield{author}{\bibinfo{person}{Qin Lin}, \bibinfo{person}{Sridha Adepu}, \bibinfo{person}{Sicco Verwer}, {and} \bibinfo{person}{Aditya Mathur}.} \bibinfo{year}{2018}\natexlab{}.
\newblock \showarticletitle{TABOR: A Graphical Model-Based Approach for Anomaly Detection in Industrial Control Systems}. In \bibinfo{booktitle}{\emph{Proceedings of the 2018 on Asia Conference on Computer and Communications Security}} (Incheon, Republic of Korea) \emph{(\bibinfo{series}{ASIACCS '18})}. \bibinfo{publisher}{Association for Computing Machinery}, \bibinfo{address}{New York, NY, USA}, \bibinfo{pages}{525–536}.
\newblock
\showISBNx{9781450355766}
\urldef\tempurl%
\url{https://doi.org/10.1145/3196494.3196546}
\showDOI{\tempurl}


\bibitem[Maci{\'{a}}-Fern{\'{a}}ndez et~al\mbox{.}(2018)]%
        {Macia-Fernandez2018_ugr}
\bibfield{author}{\bibinfo{person}{Gabriel Maci{\'{a}}-Fern{\'{a}}ndez}, \bibinfo{person}{Jos{\'{e}} Camacho}, \bibinfo{person}{Roberto Mag{\'{a}}n-Carri{\'{o}}n}, \bibinfo{person}{Pedro Garc{\'{i}}a-Teodoro}, {and} \bibinfo{person}{Roberto Ther{\'{o}}n}.} \bibinfo{year}{2018}\natexlab{}.
\newblock \showarticletitle{{UGR‘16: A new dataset for the evaluation of cyclostationarity-based network IDSs}}.
\newblock \bibinfo{journal}{\emph{Computers and Security}}  \bibinfo{volume}{73} (\bibinfo{date}{mar} \bibinfo{year}{2018}), \bibinfo{pages}{411--424}.
\newblock
\showISSN{01674048}
\urldef\tempurl%
\url{https://doi.org/10.1016/j.cose.2017.11.004}
\showDOI{\tempurl}


\bibitem[Matou{\v{s}}ek et~al\mbox{.}(2021)]%
        {Matoursek_21_automata}
\bibfield{author}{\bibinfo{person}{Petr Matou{\v{s}}ek}, \bibinfo{person}{Vojt{\v{e}}ch Havlena}, {and} \bibinfo{person}{Luk{\'a}{\v{s}} Hol{\'\i}k}.} \bibinfo{year}{2021}\natexlab{}.
\newblock \showarticletitle{Efficient modelling of ICS communication for anomaly detection using probabilistic automata}. In \bibinfo{booktitle}{\emph{2021 IFIP/IEEE International Symposium on Integrated Network Management (IM)}}. \bibinfo{publisher}{IEEE}, \bibinfo{pages}{81--89}.
\newblock


\bibitem[Mirsky et~al\mbox{.}(2018)]%
        {Mirsky_2018_kitsune}
\bibfield{author}{\bibinfo{person}{Yisroel Mirsky}, \bibinfo{person}{Tomer Doitshman}, \bibinfo{person}{Yuval Elovici}, {and} \bibinfo{person}{Asaf Shabtai}.} \bibinfo{year}{2018}\natexlab{}.
\newblock \showarticletitle{Kitsune: An Ensemble of Autoencoders for Online Network Intrusion Detection}. In \bibinfo{booktitle}{\emph{Proceedings 2018 Network and Distributed System Security Symposium}}. Internet Society.
\newblock


\bibitem[Mirsky et~al\mbox{.}(2024)]%
        {Kitsune_repo}
\bibfield{author}{\bibinfo{person}{Yisroel Mirsky}, \bibinfo{person}{Tomer Doitshman}, \bibinfo{person}{Yuval Elovici}, {and} \bibinfo{person}{Asaf Shabtai}.} \bibinfo{year}{2024}\natexlab{}.
\newblock \bibinfo{title}{Kitsune-py: A network intrusion detection system based on incremental statistics (AfterImage) and an ensemble of autoencoders (Kit{NET})}.
\newblock \bibinfo{howpublished}{\url{https://github.com/ymirsky/Kitsune-py}}.
\newblock


\bibitem[Nadeem et~al\mbox{.}(2022)]%
        {nadeem2022_alert}
\bibfield{author}{\bibinfo{person}{Azqa Nadeem}, \bibinfo{person}{Sicco Verwer}, \bibinfo{person}{Stephen Moskal}, {and} \bibinfo{person}{Shanchieh~Jay Yang}.} \bibinfo{year}{2022}\natexlab{}.
\newblock \showarticletitle{Alert-Driven Attack Graph Generation Using S-PDFA}.
\newblock \bibinfo{journal}{\emph{IEEE Transactions on Dependable and Secure Computing}} \bibinfo{volume}{19}, \bibinfo{number}{2} (\bibinfo{year}{2022}), \bibinfo{pages}{731--746}.
\newblock
\urldef\tempurl%
\url{https://doi.org/10.1109/TDSC.2021.3117348}
\showDOI{\tempurl}


\bibitem[Nadeem et~al\mbox{.}(2023)]%
        {nadeem_2023_sok}
\bibfield{author}{\bibinfo{person}{Azqa Nadeem}, \bibinfo{person}{Dani{\"e}l Vos}, \bibinfo{person}{Clinton Cao}, \bibinfo{person}{Luca Pajola}, \bibinfo{person}{Simon Dieck}, \bibinfo{person}{Robert Baumgartner}, {and} \bibinfo{person}{Sicco Verwer}.} \bibinfo{year}{2023}\natexlab{}.
\newblock \showarticletitle{Sok: Explainable machine learning for computer security applications}. In \bibinfo{booktitle}{\emph{2023 IEEE 8th European Symposium on Security and Privacy}}. IEEE, \bibinfo{pages}{221--240}.
\newblock


\bibitem[Nguyen et~al\mbox{.}(2019)]%
        {Nguyen2019_gee}
\bibfield{author}{\bibinfo{person}{Quoc~Phong Nguyen}, \bibinfo{person}{Kar~Wai Lim}, \bibinfo{person}{Dinil~Mon Divakaran}, \bibinfo{person}{Kian~Hsiang Low}, {and} \bibinfo{person}{Mun~Choon Chan}.} \bibinfo{year}{2019}\natexlab{}.
\newblock \showarticletitle{{GEE: A Gradient-based Explainable Variational Autoencoder for Network Anomaly Detection}}, In \bibinfo{booktitle}{2019 IEEE Conference on Communications and Network Security (CNS)}.
\newblock \bibinfo{journal}{\emph{2019 IEEE Conference on Communications and Network Security, CNS 2019}}, \bibinfo{pages}{91--99}.
\newblock
\showISBNx{9781538671177}
\urldef\tempurl%
\url{https://doi.org/10.1109/CNS.2019.8802833}
\showDOI{\tempurl}
\showeprint[arxiv]{1903.06661}


\bibitem[Pellegrino et~al\mbox{.}(2017)]%
        {Pellegrino2017_learning}
\bibfield{author}{\bibinfo{person}{Gaetano Pellegrino}, \bibinfo{person}{Qin Lin}, \bibinfo{person}{Christian Hammerschmidt}, {and} \bibinfo{person}{Sicco Verwer}.} \bibinfo{year}{2017}\natexlab{}.
\newblock \showarticletitle{{Learning behavioral fingerprints from Netflows using Timed Automata}}. In \bibinfo{booktitle}{\emph{Proceedings of the IM 2017 - 2017 IFIP/IEEE International Symposium on Integrated Network and Service Management}}. \bibinfo{publisher}{Institute of Electrical and Electronics Engineers Inc.}, \bibinfo{pages}{308--316}.
\newblock
\showISBNx{9783901882890}
\urldef\tempurl%
\url{https://doi.org/10.23919/INM.2017.7987293}
\showDOI{\tempurl}


\bibitem[Pennington et~al\mbox{.}(2014)]%
        {Pennington_2014_Glove}
\bibfield{author}{\bibinfo{person}{Jeffrey Pennington}, \bibinfo{person}{Richard Socher}, {and} \bibinfo{person}{Christopher~D. Manning}.} \bibinfo{year}{2014}\natexlab{}.
\newblock \showarticletitle{GloVe: Global Vectors for Word Representation}. In \bibinfo{booktitle}{\emph{Empirical Methods in Natural Language Processing (EMNLP)}}. \bibinfo{publisher}{ACL}, \bibinfo{pages}{1532--1543}.
\newblock
\urldef\tempurl%
\url{http://www.aclweb.org/anthology/D14-1162}
\showURL{%
\tempurl}


\bibitem[Sipser(2013)]%
        {sipser2013_introduction}
\bibfield{author}{\bibinfo{person}{Michael Sipser}.} \bibinfo{year}{2013}\natexlab{}.
\newblock \bibinfo{booktitle}{\emph{Introduction to the Theory of Computation} (\bibinfo{edition}{third} ed.)}.
\newblock \bibinfo{publisher}{Course Technology}, \bibinfo{address}{Boston, MA}.
\newblock
\showISBNx{113318779X}


\bibitem[Szegedy et~al\mbox{.}(2014)]%
        {szegedy2014intriguing}
\bibfield{author}{\bibinfo{person}{Christian Szegedy}, \bibinfo{person}{Wojciech Zaremba}, \bibinfo{person}{Ilya Sutskever}, \bibinfo{person}{Joan Bruna}, \bibinfo{person}{Dumitru Erhan}, \bibinfo{person}{Ian Goodfellow}, {and} \bibinfo{person}{Rob Fergus}.} \bibinfo{year}{2014}\natexlab{}.
\newblock \bibinfo{title}{Intriguing properties of neural networks}.
\newblock
\newblock
\showeprint[arxiv]{1312.6199}~[cs.CV]


\bibitem[Venturi et~al\mbox{.}(2021)]%
        {Venturi_2021_Drelab}
\bibfield{author}{\bibinfo{person}{Andrea Venturi}, \bibinfo{person}{Giovanni Apruzzese}, \bibinfo{person}{Mauro Andreolini}, \bibinfo{person}{Michele Colajanni}, {and} \bibinfo{person}{Mirco Marchetti}.} \bibinfo{year}{2021}\natexlab{}.
\newblock \showarticletitle{DReLAB - Deep REinforcement Learning Adversarial Botnet: A benchmark dataset for adversarial attacks against botnet Intrusion Detection Systems}.
\newblock \bibinfo{journal}{\emph{Data in Brief}}  \bibinfo{volume}{34} (\bibinfo{year}{2021}), \bibinfo{pages}{106631}.
\newblock
\showISSN{2352-3409}
\urldef\tempurl%
\url{https://doi.org/10.1016/j.dib.2020.106631}
\showDOI{\tempurl}


\bibitem[Verwer and Hammerschmidt(2017)]%
        {Verwer2017_Flexfringe}
\bibfield{author}{\bibinfo{person}{Sicco Verwer} {and} \bibinfo{person}{Christian~A. Hammerschmidt}.} \bibinfo{year}{2017}\natexlab{}.
\newblock \showarticletitle{{Flexfringe: A passive automaton learning package}}. In \bibinfo{booktitle}{\emph{Proceedings - 2017 IEEE International Conference on Software Maintenance and Evolution, ICSME 2017}}. \bibinfo{publisher}{IEEE}, \bibinfo{pages}{638--642}.
\newblock
\showISBNx{9781538609927}
\urldef\tempurl%
\url{https://doi.org/10.1109/ICSME.2017.58}
\showDOI{\tempurl}


\bibitem[Wong(2024)]%
        {GEE_wongmunhou}
\bibfield{author}{\bibinfo{person}{Mun~Hou Wong}.} \bibinfo{year}{2024}\natexlab{}.
\newblock \bibinfo{title}{{GEE}}.
\newblock \bibinfo{howpublished}{\url{https://github.com/munhouiani/GEE}}.
\newblock


\bibitem[Zhao(2024)]%
        {pyod_library}
\bibfield{author}{\bibinfo{person}{Yue Zhao}.} \bibinfo{year}{2024}\natexlab{}.
\newblock \bibinfo{title}{{PyOD}}.
\newblock \bibinfo{howpublished}{\url{https://pyod.readthedocs.io/en/latest/}}.
\newblock


\bibitem[Zoppi et~al\mbox{.}(2021a)]%
        {Zoppi2021_unsupervised}
\bibfield{author}{\bibinfo{person}{Tommaso Zoppi}, \bibinfo{person}{Andrea Ceccarelli}, \bibinfo{person}{Tommaso Capecchi}, {and} \bibinfo{person}{Andrea Bondavalli}.} \bibinfo{year}{2021}\natexlab{a}.
\newblock \showarticletitle{Unsupervised Anomaly Detectors to Detect Intrusions in the Current Threat Landscape}.
\newblock \bibinfo{journal}{\emph{ACM/IMS Trans. Data Sci.}} \bibinfo{volume}{2}, \bibinfo{number}{2}, Article \bibinfo{articleno}{7} (\bibinfo{date}{apr} \bibinfo{year}{2021}), \bibinfo{numpages}{26}~pages.
\newblock
\showISSN{2691-1922}
\urldef\tempurl%
\url{https://doi.org/10.1145/3441140}
\showDOI{\tempurl}


\bibitem[Zoppi et~al\mbox{.}(2021b)]%
        {Zoppi2021_meta_learning}
\bibfield{author}{\bibinfo{person}{Tommaso Zoppi}, \bibinfo{person}{Mohamad Gharib}, \bibinfo{person}{Muhammad Atif}, {and} \bibinfo{person}{Andrea Bondavalli}.} \bibinfo{year}{2021}\natexlab{b}.
\newblock \showarticletitle{Meta-Learning to Improve Unsupervised Intrusion Detection in Cyber-Physical Systems}.
\newblock \bibinfo{journal}{\emph{ACM Trans. Cyber-Phys. Syst.}} \bibinfo{volume}{5}, \bibinfo{number}{4}, Article \bibinfo{articleno}{42} (\bibinfo{date}{sep} \bibinfo{year}{2021}), \bibinfo{numpages}{27}~pages.
\newblock
\showISSN{2378-962X}
\urldef\tempurl%
\url{https://doi.org/10.1145/3467470}
\showDOI{\tempurl}


\end{thebibliography}
